%% file: final.tex
\documentclass[10pt,twocolumn,letterpaper]{article}
\usepackage{cvpr}              

\input{preamble}

\definecolor{cvprblue}{rgb}{0.21,0.49,0.74}

\def\paperID{12460} 
\def\confName{CVPR}
\def\confYear{2024}

\title{Aligning Logits Generatively for Principled Black-Box Knowledge Distillation}

\author{Jing Ma\thanks{Equal contribution, co-first author; also with Nat. Key Lab of MSIIPT.} , Xiang Xiang$^*$\thanks{Correspondence to \url{xex@hust.edu.cn}; also with Peng Cheng Lab.}\\
School of Artificial Intelligence and Automation,\\
Huazhong University of Science and Tech, Wuhan, China
\and
Ke Wang, Yuchuan Wu, Yongbin Li\\
DAMO Academy,\\
Alibaba Group, Beijing, China
}

\begin{document}
 \maketitle
 \input{sec/0_abstract}

 \input{sec/1_intro}

 \input{sec/2_relatedwork}

 \input{sec/3_method}

 \input{sec/4_experiment}

 \input{sec/5_conclusion}

{
    \small    \bibliographystyle{ieeenat_fullname}
    \bibliography{main}
}


\section*{Appendix}

In the Appendix, we provide proof of theorems and more experimental results for MEKD.
We also visualize the real and generated distributions of MEKD with DCGAN to verify the effectiveness of our method.

\section*{A. Proofs}
\label{A}

The success of deep learning can be attributed to the discovery of intrinsic structures of data, which is defined as the manifold distribution hypothesis \cite{tenenbaum2000global}. The data is concentrated on a manifold $\Sigma\in \mathbb{R}^n$, which is embedded in the image space $\mathcal{X}$, and data distribution can be abstracted as a probability distribution $\mu$ over the data manifold. The encoding-map $\varphi:\Sigma\rightarrow\Omega$ maps the data manifold $\Sigma$ to the label manifold $\Omega\in \mathbb{R}^C$ in a label space $\mathcal{Y}$ which is also called latent space, while mapping the data distribution $\mu$ to latent distribution $\upsilon=\varphi_\#\mu$. Each sample $x$ is mapped from the image space into the latent space, and its result $\varphi(x)$ is called a latent code. The decoding-map $\varphi^{-1}$ remaps latent codes to the data manifold. Both $\varphi$ and $\varphi^{-1}$ are strongly nonlinear functions, which can be simulated with different neural networks \cite{lei2019geometric,lei2020geometric}.
Meanwhile, the well-known Kolmogorov Theorem \cite{koppen2002training,braun2009constructive} indicates that any multivariate continuous function can be represented as the sum of continuous real-valued functions with continuous one-dimensional outer and inner functions $\Phi_q$ and $\Psi_{q,p}$.

The teacher function $f_T\in\varphi$ can be considered as a kind of encoding map, and the generator function $f_G\in\varphi^{-1}$ can be considered as a kind of decoding map.
Let $\mathcal{X}\in \mathbb{R}^n$ be the image space, where data $x$ is sampled from. For a $C$-way classification task, let $\mathcal{Y}\in\mathbb{R}^C$ be the latent space, where $|\mathcal{Y}|=C$. Defining the model as a complex mapping function from the image distribution to the latent distribution, we can consider the teacher model as $f_T:\mathcal{X}\rightarrow\mathcal{Y}$ parameterized by $\theta_T\in\Theta_T$, whose outputs indicate the probabilities (\emph{e.g.}, logits) of what category the samples belong to. The same for the student model $f_S:\mathcal{X}\rightarrow\mathcal{Y}$ parameterized by $\theta_S\in\Theta_S$. 


\begin{definition}
\label{definition1} {\bf(Function Equivalence)}
Giving the student and teacher model $f_S$ and $f_T$, for a data distribution $\mu\in\mathcal{X}$ in image space which is mapped to $\mathbb{P}_S\in\mathcal{Y}$ and $\mathbb{P}_T\in\mathcal{Y}$ in latent space. If the Wasserstein distance between $\mathbb{P}_S$ and $\mathbb{P}_T$ equals zero,

\begin{equation}
\label{W}
    W(\mathbb{P}_S, \mathbb{P}_T)=\inf_{\gamma\in\Pi(\mathbb{P}_S, \mathbb{P}_T)}\mathbb{E}_{(y_S,y_T)\sim\gamma}\left[\ \|y_S-y_T\| \ \right] = 0,
\end{equation}

\noindent the student and teacher model are equivalent, \emph{i.e.}, $f_S=f_T$, where $\Pi(\mathbb{P}_S,\mathbb{P}_T)$ is the set of all joint distributions $\gamma(y_S,y_T)$ whose marginals are $\mathbb{P}_S$ and $\mathbb{P}_T$, respectively.
\end{definition}

\begin{definition}
\label{definition2} {\bf(Inverse Mapping)}
Giving a prior distribution $p\in\mathbb{R}^C$, for a data distribution $\mu\in\mathbb{R}^n$, if the Wasserstein distance between generated distribution $\mu'=(f_G)_{\#}p$ and $\mu$ equals zero,

\begin{equation}
    W(\mu', \mu)=\inf_{\gamma\in\Pi(\mu',\mu)}\mathbb{E}_{(x',x)\sim\gamma}[\ \|x'-x\|\ ]=0,
\end{equation}

\noindent then the generator $f_G:\mathbb{R}^C\rightarrow\mathbb{R}^n$ is the inverse mapping of the teacher function $f_T:\mathbb{R}^n\rightarrow\mathbb{R}^C$, denoted as $f_G=f_T^{-1}$, where $\Pi(\mu',\mu)$ is the set of all joint distributions $\gamma(x',x)$ whose marginals are respectively $\mu'$ and $\mu$.
\end{definition}

\subsection*{A.1. Proof of Theorem 1}

\begin{theorem}
\label{theorem1} {\bf(Empirical Approximation)}
For any $0<\epsilon<1/2$ and any integer $m>4$, let $g:\mathbb{R}^C\rightarrow\mathbb{R}^n$ be the mapping function of generator $G$ with $n\leq\frac{20\log m}{\epsilon^2}$. For two sets $V_S=\{y_S:y_S\in \mathbb{P}_S\}$ and $V_T=\{y_T:y_T\in \mathbb{P}_T\}$, both of which have $m$ points in $\mathbb{R}^C$, if the empirical Wasserstein distance between $g(V_S)$ and $g(V_T)$ equals zero,

\begin{equation}
    \hat{W}(g(V_S),g(V_T))=\frac{1}{m}\sum_{i=1}^m\|g(y_S^i)-g(y_T^i)\|=0,
\end{equation}

\noindent then $W(\mathbb{P}_S, \mathbb{P}_T)=0$.
\end{theorem}

\begin{proof}
According to Johnson-Lindenstrauss theorem, for $y_S\in V_S$ and $y_T\in V_T$, we have
\begin{align}
    \|y_S-y_T\|\le(1+\epsilon)\|g(y_S)-g(y_T)\|.
\end{align}

For set $V_S$ and $V_T$, we can get the empirical Wasserstein distance between them:
\begin{align}
\begin{aligned}
\hat{W}(V_S,V_T)
&=\frac{1}{m}\sum_{i=1}^m\|y_S^i-y_T^i\|\\
&\le \frac{1}{m}\sum_{i=1}^m(1+\epsilon)\|g(y_S^i)-g(y_T^i)\|\\
&= \frac{1+\epsilon}{m}\sum_{i=1}^m\|g(y_S^i)-g(y_T^i)\|\\
&= (1+\epsilon)\hat{W}(g(V_S),g(V_T)) = 0.
\end{aligned}
\end{align}

Because the Wasserstein distance between $\mathbb{P}_S$ and $\mathbb{P}_T$ is the expectation of the empirical Wasserstein distance between $V_S$ and $V_T$, i.e.,
\begin{align}
W(\mathbb{P}_S, \mathbb{P}_T) = \mathbb{E}_{(V_S,V_T)\sim\Pi(\mathbb{P}_S,\mathbb{P}_T)}\left[\hat{W}(V_S,V_T)\right],
\end{align}
so we can get
\begin{align}
W(\mathbb{P}_S, \mathbb{P}_T)\le \hat{W}(V_S,V_T)=0.
\end{align}
Since 
\begin{align}
W(\mathbb{P}_S, \mathbb{P}_T)=\inf_{\gamma\in\Pi(\mathbb{P}_S, \mathbb{P}_T)}\mathbb{E}_{(y_S,y_T)\sim\gamma}\left[\ \|y_S-y_T\| \ \right] \ge 0,
\end{align} 
then the result $W(\mathbb{P}_S, \mathbb{P}_T)=0$ is derived.
\end{proof}

\subsection*{A.2. Proof of Theorem 2}

\begin{theorem}
\label{theorem2} {\bf(Optimization Direction)}
Let $\mu\in\mathcal{X}$ be any distribution. $f_S, f_T, f_G$ are the mapping functions of the student, teacher, and generator, respectively. $f_S$ is parameterized by $\theta_S\in\Theta_S$. Then, when 
\begin{equation}
    \min_{\theta_S\in\Theta_S} \mathbb{E}_{x\sim \mu} \left[\|f_G\circ f_S(x), f_G\circ f_T(x)\|\right]\rightarrow 0,
\end{equation}
it holds that $f_S\rightarrow f_T$, and we have
\begin{align}
    &\nabla_{\theta_S}\mathbb{E}_{x\sim\mu}[f_S(x)]=\nabla_{\theta_S}W(\mathbb{P}_S, \mathbb{P}_T)\nonumber \\
    &\ \ \ \ \ \ \ \ =\mathbb{E}_{x\sim\mu}[\nabla_{\theta_S}\|f_G\circ f_S(x)-f_G\circ f_T(x)\|].
\end{align}
\end{theorem}

\begin{proof}
Let us define
\begin{align}
\label{V}
V(f_S,\theta_S)=\mathbb{E}_{x\sim\mu}\left[\ \|f_S(x),f_T(x)\|\ \right],
\end{align}
\begin{align}
\label{V_}
V'(f_S, \theta_S)=\mathbb{E}_{x\sim\mu}[\ \|f_G\circ f_S(x), f_G\circ f_T(x)\|\ ],
\end{align}
\noindent where $f_S$ lies in $\mathcal{F_S}=\{f_S:\mathcal{X}\rightarrow\mathcal{Y}\}$ and $\theta_S\in\Theta_S$.

According to the Johnson-Lindenstrauss Lemma \cite{frankl1988johnson}, for any $0<\epsilon<1/2$ and any integer $m>4$, let $n= \frac{20\log m}{\epsilon^2}$, then for any set $S$ of $m$ points in $\mathbb{R}^C$, the generator mapping function $f_G:\mathbb{R}^C\rightarrow\mathbb{R}^n$  for all $f_S(x),f_T(x)\in S$ holds that
\begin{align}
    (1-\epsilon)\ \|f_G\circ f_S(x), f_G\circ f_T(x)\| \ \ \ \ \ \ \ \ \ \ \ \ \ \ \ \ \ \nonumber\\
    \le \|f_S(x),f_T(x)\| \ \ \ \ \ \ \ \ \ \ \ \ \ \ \ \ \ \ \ \ \ \ \ \ \ \ \ \ \ \ \ \ \ \ \  \nonumber\\
    \le (1+\epsilon)\ \|f_G\circ f_S(x), f_G\circ f_T(x)\|.
\end{align}

Using Squeeze Theorem \cite{lewis2000nonsmooth}, we know that the minimization of equation \ref{V} and equation \ref{V_} converge to the same results, \emph{i.e.},
\begin{equation}
    \inf V(f_S,\theta_S) = \inf V'(f_S,\theta_S).
\end{equation}
We can rewrite the equation \ref{W} using $x\sim\mu$:

\begin{equation}
\begin{aligned}
W(\mathbb{P}_S,\mathbb{P}_T)&=\inf_{\gamma\in{\prod}(\mathbb{P}_S, \mathbb{P}_T)}\mathbb{E}_{(y_S,y_T)\sim\gamma}\left[\ \|y_S-y_T\| \ \right]\\
&=\inf_{\gamma\in\prod(f_S(\mu),f_T(\mu))}\mathbb{E}_{x\sim\mu}\left[\ \|f_S(x),f_T(x)\|\ \right]\\
&=\inf_{\gamma\in\prod(f_S(\mu),f_T(\mu))}V(f_S,\theta_S),
\end{aligned}
\end{equation}
where $f_S$ and $f_T$ map distribution $\mu$ to $\mathbb{P}_S$ and $\mathbb{P}_T$, respectively. So we can get 
\begin{equation}
    \inf\ V'(f_S,\theta_S)=\inf\ V(f_S,\theta_S)=W(\mathbb{P}_S, \mathbb{P}_T).
\end{equation}

According to Def. \ref{definition1}, when $\inf V'(f_S,\theta_S)\rightarrow 0$, then $W(\mathbb{P}_S,\mathbb{P}_T) \rightarrow 0$, and we can derive that $f_S\rightarrow f_T$. 

The rest of the proof will be dedicated to show that the optimal solution of $\min V'(f_S,\theta_S)$ leads to reduce the Wasserstein distance of $\mathbb{P}_S$ and $\mathbb{P}_T$, which drives $f_S$ to approximate $f_T$. 

We know by the Kantorovich-Rubinstein duality \cite{villani2009optimal} that there is an $\tilde{f}_S\in\mathcal{F}_S$ that attains 

\begin{align}
    \inf\ \mathbb{E}_{x\sim\mu}[\ \|\tilde{f}_S(x),f_T(x)\|\ ]\ \ \ \ \ \ \ \ \ \ \ \ \ \ \ \ \ \ \ \ \ \ \ \ \ \ \ \ \ \ \nonumber\\
    =\sup\ \mathbb{E}_{x\sim\mu}[\ \tilde{f}_S(x)\ ]-\mathbb{E}_{x\sim\mu}[\ f_T(x)\ ].
\end{align}

Let us define $\tilde{X}(\theta_S)=\{\tilde{f}_S\in\mathcal{F}_S:V(\tilde{f}_S,\theta_S)=W(\mathbb{P}_S,\mathbb{P}_T)\}$ which is non-empty. We know by a simple envelope theorem \cite{milgrom2002envelope} that 
\begin{equation}
\begin{aligned}
\nabla_{\theta_S}W(\mathbb{P}_S,\mathbb{P}_T)=\nabla_{\theta_S}V(\tilde{f}_S,\theta_S),
\end{aligned}
\end{equation}
for any $\tilde{f}_S\in \tilde{X}(\theta_S)$ when both terms are well-defined. 

Let $\tilde{f}_S\in \tilde{X}(\theta_S)$, which we knows exists since $\tilde{X}(\theta_S)$ is non-empty for all $\theta_S$. Then, we get

\begin{equation}
\begin{aligned}
\nabla_{\theta_S}W(\mathbb{P}_S,\mathbb{P}_T)&=\nabla_{\theta_S}V(\tilde{f}_S,\theta_S)\\
&=\nabla_{\theta_S}\mathbb{E}_{x\sim\mu}[\ \|\tilde{f}_S(x),f_T(x)\|\ ]\\
&=\nabla_{\theta_S}\mathbb{E}_{x\sim\mu}[\ \tilde{f}_S(x)\ ]-\mathbb{E}_{x\sim\mu}\left[\ f_T(x)\ \right]\\
&=\nabla_{\theta_S}\mathbb{E}_{x\sim\mu}[\ \tilde{f}_S(x)\ ].
\end{aligned}
\end{equation}

In practice, we use empirical distance
between generated images of the student and teacher as loss to update $\theta_S$ by back-propagation, \emph{i.e.},
\begin{align}
    &\nabla_{\theta_S}\mathbb{E}_{x\sim\mu}[f_S(x)]=\nabla_{\theta_S}W(\mathbb{P}_S, \mathbb{P}_T)\nonumber \\
    &\ \ \ \ \ \ \ \ =\nabla_{\theta_S}W((f_G)_{\#}\mathbb{P}_S,(f_G)_{\#}\mathbb{P}_T)\nonumber\\
    &\ \ \ \ \ \ \ \ =\nabla_{\theta_S}\mathbb{E}_{x\sim \mu}[\|f_G\circ f_S(x)-f_G\circ f_T(x)\|]\nonumber\\
    &\ \ \ \ \ \ \ \ =\mathbb{E}_{x\sim\mu}[\nabla_{\theta_S}\|f_G\circ f_S(x)-f_G\circ f_T(x)\|],
\end{align}
\noindent when $W(\mathbb{P}_S,\mathbb{P}_T)\rightarrow 0$, the student function $f_S$ converges to the teacher function $f_T$.
\end{proof}

\subsection*{A.3. Proof of Theorem 3}

\begin{theorem}{\bf(Generalization Bound)}
\label{theorem3}
Let $H\subseteq \mathbb{R}^{\mathcal{X}\times\mathcal{Y}}$ be a hypothesis set for $C$-way classification task. For any $0<\epsilon<1/2$ and a sample $S$ of size $m>4$ drawn according to $\mu$, let $g:\mathbb{R}^C\rightarrow\mathbb{R}^n$ be a mapping function of generator $G$ with $n\leq\frac{20\log m}{\epsilon^2}$. Fix $\rho>0$, for any $1>\delta>0$, with probability at least $1-\delta$, the following holds for all $h\in H$,
\begin{equation}
    R(h)\le \hat{R}_{\rho}(h)+\frac{2C^2}{\rho(1-\epsilon)}\sqrt{\frac{r^2\Lambda^2}{m}}+\sqrt{\frac{\log\frac{1}{\delta}}{2m}}.
\end{equation}
For any $x\in\mathcal{X}$, the $\Lambda \geq 0$ and $(\sum_{y=1}^{C} \|h(x,y)\|^p)^{1/p}\leq \Lambda$ for any $p \geq 1$, and the $r>0$ for $K(x,x)\leq r^2$ where kernel $K:\mathcal{X}\times\mathcal{X}\rightarrow\mathbb{R}$ is positive definite symmetric.
\end{theorem}

\begin{proof}
For the $C$-way classification task, a hypothesis $h:\mathcal{X}\times\mathcal{Y}\rightarrow\mathbb{R}$ aims to get $y$ with the minimum distance, i.e. $\arg\min_{y\in\mathcal{Y}}\|\overline{h}(x)-\overline{h}_y\|$ which is equivalent to $\arg\min_{y\in\mathcal{Y}}(1+\epsilon)\|g(\overline{h}(x))-g(\overline{h}_y)\|$ by Johnson-Lindenstrauss theorem, as the result of $x$. We define the margin $\rho_h(x,y)$ of the hypothesis $h$ as
\begin{align}
\rho_h(x,y)=\|g(\overline{h}(x))-g(\overline{h}_{y})\|-\min_{y'\ne y}\|g(\overline{h}(x))-g(\overline{h}_{y'})\|,
\end{align}
where $\overline{h}(x)$ is the vector of $h(x,y),y\in\mathcal{Y}$ and $\overline{h}_{y}$ use the mean of $x$ which belong to class $y$ as input. $g$ is the mapping function of generator $G$.

For any $\rho<0$, we can define the empirical margin loss of hypothesis $h$ for multi-class classification as
\begin{align}
\hat{R}_{\rho}(h)=\frac{1}{m}\sum_{i=1}^m\Phi_{\rho}(\rho_{h}(x_i,y_i)),
\end{align}
where $\Phi_{\rho}$ is the margin loss function
\begin{align}
\Phi_{\rho}(x)=
\left\{
\begin{array}{l}
1\ \ \ \ \ \ \ \ \ \ \ \ \ \ \ \ 0\le x,\\
1-x/\rho\ \ \ \ \ \rho\le x\le0,\\
0\ \ \ \ \ \ \ \ \ \ \ \ \ \ \ \ x\le\rho.
\end{array}
\right.
\end{align}
Thus, empirical margin loss is upper bounded by
\begin{align}
\hat{R}_{\rho}(h)\le \frac{1}{m}\sum_{i=1}^m \mathbbm{1}_{\rho_h(x_i,y_i)\ge\rho}.
\end{align}

Let $\tilde{H}=\{(x,y)\mapsto \rho_h(x,y):h\in H\}$, consider the family of functions $\tilde{\mathcal{H}}=\{\Phi_{\rho}\circ r:r\in \tilde{H}\}$ derived from $\tilde{H}$, which take values in $[0,1]$. By Rademacher theorem, with the probability at least $1-\delta$, for all $h\in H$,
\begin{align}
E[\Phi_{\rho}(\rho_h(x,y))]\le \hat{R}_{\rho}(h)+2\mathcal{R}_m(\Phi\circ\hat{H})+\sqrt{\frac{\log\frac{1}{\delta}}{2m}}.
\end{align}

Since $\mathbbm{1}_{\mu\ge0}\le\Phi_{\rho}(\mu)$ for all $\mu\in\mathbb{R}$, the generalization error $R(h)$ is a lower bound on the left-hand side by Johnson-Lindenstrauss theorem, $R(h)=E\left[\mathbbm{1}_{\|\overline{h}(x)-\overline{h}_{y}\|-\min_{y'\ne y}\|\overline{h}(x)-\overline{h}_{y'}\|\ge 0}\right]\le E[\Phi_{\rho}(\rho_h(x,y))]$, and we get
\begin{align}
R(h)\le \hat{R}_{\rho}(h)+2\mathcal{R}_m(\Phi\circ\hat{H})+\sqrt{\frac{\log\frac{1}{\delta}}{2m}}.
\end{align}

Let $\rho=-\rho$, because the $(1/\rho)$-Lipschitzness of $\Phi_p$, so that $\mathcal{R}_m(\Phi_p\circ \tilde{H})\le \frac{1}{\rho}\mathcal{R}_m(\tilde{H})$. Here, $\mathcal{R}_m(\tilde{H})$ can be upper bounded as follows:
\begin{align}
\begin{aligned}
\mathcal{R}_m(\tilde{H})&=\frac{1}{m}\mathop{E}_{S,\sigma}[\sup_{h\in H}\sum_{i=1}^m\sigma_i\rho_h(x_i,y_i)]\\
&=\frac{1}{m}\mathop{E}_{S,\sigma}[\sup_{h\in H}\sum_{i=1}^m\sum_{y\in \mathcal{Y}}\sigma_i\rho_h(x_i,y)\mathbbm{1}_{y=y_i}]\\
&\le\frac{1}{m}\sum_{y\in \mathcal{Y}}\mathop{E}_{S,\sigma}[\sup_{h\in H}\sum_{i=1}^m\sigma_i\rho_h(x_i,y)\mathbbm{1}_{y=y_i}]\\
&=\frac{1}{m}\sum_{y\in \mathcal{Y}}\mathop{E}_{S,\sigma}[\sup_{h\in H}\sum_{i=1}^m\sigma_i\rho_h(x_i,y)(\frac{2(\mathbbm{1}_{y=y_i})-1}{2}+\frac{1}{2})]\\
&\le \frac{1}{2m}\sum_{y\in\mathcal{Y}}\mathop{E}_{S,\sigma}[\sup_{h\in H}\sum_{i=1}^m\sigma_i(2(\mathbbm{1}_{y=y_i})-1)\rho_h(x_i,y)]+\\
&\ \ \ \ \ \ \ \ \ \ \ \ \ \frac{1}{2m}\sum_{y\in \mathcal{Y}}\mathop{E}_{S,\sigma}[\sup_{h\in H}\sum_{i=1}^m\sigma_i\rho_h(x_i,y)]\\
&=\frac{1}{m}\sum_{y\in\mathcal{Y}}\mathop{E}_{S,\sigma}[\sup_{h\in H}\sum_{i=1}^m\sigma_i\rho_h(x_i,y)],
\end{aligned}
\end{align}
where $\mathbf{\sigma}=(\sigma_1,\ldots,\sigma_m)^T$ with $\sigma_i$ independent uniform random variables taking values in $\{-1,+1\}$, observing that $\sigma_i$ and $-\sigma_i$ are distributed in the same way.

Let $\Pi_1(H)^{(C-1)}=\{\min\{h_1,\ldots,h_l\}:h_i\in\Pi_1(H),i\in[1,C-1]\}$. By Johnson-Lindenstrauss theorem, we get
\begin{align}
\begin{aligned}
\mathcal{R}_m(\tilde{H})
&\le \frac{1}{m}\sum_{y\in\mathcal{Y}}\mathop{E}_{S,\sigma}[\sup_{h\in H}\sum_{i=1}^m \sigma_i(\|g(\overline{h}(x))-g(\overline{h}_{y})\| \\
&\ \ \ \ \ \ \ \ \ \ \ \ \ -\min_{y'\ne y}\|g(\overline{h}(x))-g(\overline{h}_{y'})\|)] \\
&\le \frac{1}{m}\sum_{y\in\mathcal{Y}}\mathop{E}_{S,\sigma}[\sup_{h\in H}\sum_{i=1}^m \sigma_i\frac{1}{1-\epsilon}(\|\overline{h}(x_i)\\
&\ \ \ \ \ \ \ \ \ \ \ \ \ -\overline{h}_y\|-\min_{y'\ne y}\|\overline{h}(x_i)-\overline{h}_{y'}\|)]\\
&\le \frac{1}{(1-\epsilon)m}\sum_{y\in\mathcal{Y}}[\mathop{E}_{S,\sigma}[\sup_{h\in H}\sum_{i=1}^m\sigma_i\|\overline{h}(x_i)-\overline{h}_y\|]\\
&\ \ \ \ \ \ \ \ \ \ \ \ \ +\mathop{E}_{S,\sigma}[\sup_{h\in H}\sum_{i=1}^m\sigma_i\min_{y'\ne y}\|\overline{h}(x_i)-\overline{h}_{y'}\|]]\\
&\le \frac{1}{(1-\epsilon)m}\sum_{y\in\mathcal{Y}}[\mathop{E}_{S,\sigma}[\sup_{h\in \Pi_1(H)}\sum_{i=1}^m\sigma_ih(x_i)]\\
&\ \ \ \ \ \ \ \ \ \ \ \ \ +\mathop{E}_{S,\sigma}[\sup_{h\in \Pi_1(H)^{(C-1)}}\sum_{i=1}^m\sigma_ih(x_i)]]\\
&\le \frac{C}{(1-\epsilon)m}[C\mathop{E}_{S,\sigma}[\sup_{h\in\Pi_1(H)}\sum_{i=1}^m\sigma_ih(x_i)]]\\
&= \frac{C^2}{1-\epsilon}[\frac{1}{m}\mathop{E}_{S,\sigma}[\sup_{h\in\Pi_1(H)}\sum_{i=1}^m\sigma_ih(x_i)]]\\
&= \frac{C^2}{1-\epsilon}\mathcal{R}_m(\Pi_1(H)).
\end{aligned}
\end{align}

Let $K:\mathcal{X}\times\mathcal{X}\rightarrow\mathbb{R}$ be a positive definite symmetric kernel and let $h(x,y)= \arg\max_{y\in\mathcal{Y}} w_y\cdot \Phi(x)$, where $\Phi:\mathcal{X}\rightarrow \mathbb{R}^n$ be a feature mapping associated to $K$. We denote $W$ as $W=(w_1^{\top},\ldots,w_C^{\top})$. For any $p\ge 1$, the family of kernel-based hypotheses is 
\begin{align}
    H=\{h\in\mathcal{R}^{\mathcal{X}\times\mathcal{Y}}:h(x,y)\in\mathbb{R}^n,\|h\|_p\le \Lambda\},
\end{align}
where $\|h\|_p=(\sum_{y=1}^C \|h(x,y)\|^p)^{1/p}$.

Observe that for all $l\in[1,C]$, we have $\|w_l\|\le (\sum_{l=1}^C \|w_l\|^p)^{1/p}=\|W\|_p\le\|h\|_p\le\Lambda$.And for $i\ne j$, $\mathop{E}_{\sigma}[\sigma_i, \sigma_j]=0$. The Radmacher complexity of the hypotheses set $\Pi_1(H)$ can be expressed and bounded as follows:

\begin{align}
    \mathcal{R}_m(\Pi_1(H))&=\frac{1}{m}\mathop{E}_{S,\sigma}\left[\sup_{y\in\mathcal{Y},\|W\|\le\Lambda}\left\langle w_y, \sum_{i=1}^m \sigma_i\Phi(x_i) \right\rangle\right] \nonumber\\
    &\le\frac{1}{m}\mathop{E}_{S,\sigma}\left[ \sup_{y\in\mathcal{Y},\|W\|\le\Lambda}\|w_y\|\left\|\sum_{i=1}^m\sigma_i\Phi(x_i)\right\|\right] \nonumber \\
    &\le \frac{\Lambda}{m}\mathop{E}_{S,\sigma}\left[\left\|\sum_{i=1}^m\sigma_i\Phi(x_i)\right\|\right] \nonumber\\
    &\le \frac{\Lambda}{m} \left[ \mathop{E}_{S,\sigma}\left[ \left\|\sum_{i=1}^m\sigma_i\Phi(x_i)\right\|^2 \right] \right]^{1/2} \nonumber \\
    &=\frac{\Lambda}{m}\left[ \mathop{E}_{S,\sigma}\left[ \sum_{i=1}^m \|\Phi(x_i)\|^2 \right] \right]^{1/2} \nonumber \\
    &=\frac{\Lambda}{m}\left[ \mathop{E}_{S,\sigma} \left[\sum_{i=1}^m K(x_i,x_i)\right] \right]^{1/2} \nonumber \\
    &\le\frac{\Lambda\sqrt{mr^2}}{m}=\sqrt{\frac{r^2\Lambda^2}{m}},
\end{align}
which concludes the proof.

\end{proof}

\setlength{\tabcolsep}{4pt}
\renewcommand\arraystretch{1.2}
\begin{table*}[t]
\vspace{3mm}
\begin{center}
\setlength{\tabcolsep}{1.2mm}{
\begin{tabular}{cc|cccc|cccc}
    \toprule
    Method  & Data Size & \multicolumn{4}{c|}{MNIST}  & \multicolumn{4}{c}{CIFAR-10} \\
    \hline
    \multirow{2}{*}{Teacher} & \multirow{2}{*}{50K$\sim$100K} & ResNet32 & VGG13  & ResNet32 & ResNet32 & ResNet56 & VGG13 & ResNet56 & ResNet56\\
    & & 99.50 & 99.52 & 99.50 & 99.50 & 94.15 & 94.42  & 94.15 & 94.15\\
    \multirow{2}{*}{Student} & \multirow{2}{*}{50K$\sim$100K} & ResNet8 & VGG11 & VGG11 & MobileNet & ResNet8 & VGG11 & VGG11 & MobileNet\\ 
    & & 99.24 & 99.41 & 99.41 & 99.18 & 87.74 & 91.81 & 91.81 & 90.04  \\
	\hline
	 KD \cite{hinton2015distilling} & 50K$\sim$100K & 99.33 & 99.44 & 99.31 & 99.30 & 86.58 & 92.16 & 92.25 & 90.43 \\
	 ML \cite{ba2014deep} & 50K$\sim$100K & 99.49 & 99.40 & 99.44 & 99.40 & 87.89 & 91.58 & 91.91 & 91.19 \\
	 AL \cite{wang2018adversarial} & 50K$\sim$100K & 99.37 & 99.26 & 99.26 & 99.21 & 87.25 & 91.96 & 91.97 & 90.54 \\
     DKD \cite{zhao2022decoupled} & 50K$\sim$100K & 99.33 & 99.43 & 99.48 & 99.42 & 86.61 & 92.06 & 92.42 & 90.50 \\
     DAFL \cite{chen2019data} & 0K & 96.42 & 97.00 & 96.14 & 97.85 & 60.67 & 65.41 & 66.03 & 69.59 \\
     \hline
     KN \cite{orekondy2019knockoff} & 10K & 98.61 & 98.81 & 98.07 & 98.54 & 80.62 & 81.83 & 82.41 & 85.07 \\
     AM \cite{wang2020neural} & 10K & 99.33 & 99.47 & 99.50 & 99.42 & 74.89 & 77.25 & 74.26 & 73.65 \\
     DB3KD \cite{wang2021zero} & 10K & 98.94 & 99.16 & 98.91 & 98.91 & 78.47 & 83.72 & 85.84 & 81.67 \\
	 MEKD (soft)  & 10K & 99.40 & 99.43 & 99.36 & 99.25 & 85.36 & 86.11 & 87.27 & 86.85 \\
    MEKD (hard) & 10K & 99.40 & 99.45 & 99.28 & 99.27 & 84.45 & 86.16 & 87.25 & 86.53 \\
    \bottomrule
\end{tabular}}
\vspace{0mm}
\caption{Top-1 classification accuracy (\%) of the student model on MNIST and CIFAR-10.}
\label{table:table1}
\end{center}
\end{table*}
\setlength{\tabcolsep}{1.4pt}

\setlength{\tabcolsep}{4pt}
\renewcommand\arraystretch{1.2}
\begin{table*}[t]
\begin{center}
\setlength{\tabcolsep}{1.2mm}{
\begin{tabular}{cc|cccc|cccc}
    \toprule
    Method  & Data Size & \multicolumn{4}{c|}{CIFAR-100} & \multicolumn{4}{c}{Tiny ImageNet}\\
    \hline
    \multirow{2}{*}{Teacher} & \multirow{2}{*}{50K$\sim$100K} & ResNet56 & VGG13  & ResNet56 & ResNet56 & ResNet110 & VGG13 & ResNet110 & ResNet110\\
    & & 72.06 & 74.68 & 72.06 & 72.06 & 60.71 & 59.89 & 60.71 & 60.71 \\
    \multirow{2}{*}{Student} & \multirow{2}{*}{50K$\sim$100K} & ResNet8 & VGG11 & VGG11 & MobileNet & ResNet32 & VGG11 & VGG11 & MobileNet\\ 
    & & 59.92 & 69.12 & 69.12 & 68.14 & 55.47 & 54.14 & 54.14 & 56.07 \\
	\hline
	 KD \cite{hinton2015distilling} & 50K$\sim$100K & 53.31 & 70.88 & 67.97 & 71.86 & 54.14 & 54.40 & 49.63 & 57.85 \\
	 ML \cite{ba2014deep} & 50K$\sim$100K & 54.44 & 67.78 & 70.18 & 73.08 & 56.56 & 57.46 & 56.78 & 60.07 \\
	 AL \cite{wang2018adversarial} & 50K$\sim$100K & 58.36 & 69.92 & 71.13 & 71.33 & 46.02 & 46.26 & 45.60 & 51.29 \\
     DKD \cite{zhao2022decoupled} & 50K$\sim$100K & 54.28 & 67.32 & 70.10 & 72.38 & 55.99 & 55.88 & 56.52 & 59.43 \\
     DAFL \cite{chen2019data} & 0K & 42.44 & 43.78 & 48.32 & 54.10 & 38.44 & 31.93 & 34.13 & 40.93 \\
     \hline
     KN \cite{orekondy2019knockoff} & 10K & 48.75 & 57.83 & 55.64 & 58.49 & 48.92 & 46.99 & 45.05 & 50.22 \\
     AM \cite{wang2020neural} & 10K & 50.69 & 62.17 & 63.20 & 65.58 & 47.72 & 49.26 & 47.32 & 51.54 \\
     DB3KD \cite{wang2021zero} & 10K & 50.49 & 63.48 & 62.76 & 63.67 & 47.95 & 48.46 & 46.93 & 50.49 \\
	 MEKD (soft)  & 10K & 51.87 & 64.76 & 64.83 & 67.07 & 50.87 & 51.85 & 49.95 & 54.93 \\
    MEKD (hard) & 10K & 51.67 & 64.72 & 65.32 & 67.36 & 49.89 & 51.33 & 49.36 & 54.71 \\
    \bottomrule
\end{tabular}}
\vspace{0mm}
\caption{Top-1 classification accuracy (\%) of the student model on CIFAR-100 and Tiny ImageNet.}
\label{table:table2}
\end{center}
\vspace{-5mm}
\end{table*}
\setlength{\tabcolsep}{1.4pt}

\setlength{\tabcolsep}{4pt}
\renewcommand\arraystretch{1.1}
\begin{table*}[th!]
\begin{center}
\setlength{\tabcolsep}{1.7mm}{
\scalebox{.5}{}
\begin{tabular}{c|c|c|c|c|c|c|c|c|c|c|c}
    \toprule
    \multirow{2}{*}{Dataset} & T - S & \multirow{2}{*}{Data Size} & KD & ML & AL & DKD & KN & AM & DB3KD & MEKD & MEKD \\
    & Pairs & & (soft) & (soft) & (soft) & (soft) & (soft) & (soft) & (hard) & (soft) & (hard)\\
    \midrule
    \multirow{2}{*}{ImageNet-1K} & RN50 - RN34 & 100K & 52.08 & 54.97 & 53.50 & 53.57 & 56.77 & 56.92 & 58.61 & 59.89 & 59.32 \\
    & RX101 - RX50 & 100K & 54.90 & 56.58 & 50.88 & 55.31 & 57.43 & 55.64 & 59.90 & 61.21 & 60.54 \\
    \bottomrule
\end{tabular}}
\vspace{1mm}
\caption{
Top-1 classification accuracy (\%) of the student model on ImageNet-1K. 
We use pretrained RN50 ($76.13\%$) and RX101 ($79.31\%$) as the teacher models, respectively. RN is ResNet and RX is ResNeXt}
\label{table:table3}
\end{center}
\end{table*}
\setlength{\tabcolsep}{1.4pt}

\setlength{\tabcolsep}{4pt}
\renewcommand\arraystretch{1.2}
\begin{table*}[t]
\vspace{1mm}
\begin{center}
\setlength{\tabcolsep}{1.9mm}{
\scalebox{.5}{}
\begin{tabular}{c|c|c|c|c|c|c|c|c|c|c|c}
    \toprule
    \multirow{2}{*}{Dataset} & T - S & \multirow{2}{*}{Data Size} & KD & ML  & AL & DKD & KN & AM & DB3KD & MEKD & MEKD \\
    & Pairs & & (soft) & (soft) & (soft) & (soft) & (soft) & (soft) & (hard) & (soft) & (hard)\\
    \midrule
    \multirow{4}{*}{CIFAR10} & T: ResNet56 & 0.1K & 16.74 & 17.78 & 12.97 & 20.66 & 27.67 & 48.31 & 43.05 & 49.04 & 47.12 \\
    & (94.15\%) & 1K & 31.25 & 31.57 & 32.05 & 31.09 & 58.65 & 62.05 & 64.28 & 69.84 & 68.66 \\
    & S: MobileNet & 10K & 70.90 & 73.06 & 68.61 & 75.44 & 85.07 & 73.65 & 81.67 & 86.85 & 86.53 \\
    & (90.04\%) & 50K(full) & 90.43 & 91.19 & 90.54 & 90.50 & 92.19 & 86.33 & 92.46 & 93.48 & 93.09 \\
    \midrule
    \multirow{4}{*}{CIFAR100} & T: ResNet56 & 0.1K  & 01.96 & 01.88 & 01.72 & 02.56 & 13.23 & 36.73 & 30.72 & 33.56 & 34.60 \\
    & (72.06\%) & 1K & 10.36 & 10.06 & 09.62 & 10.81 & 35.80 & 52.09 & 50.14 & 53.84 & 54.52 \\
    & S: MobileNet & 10K & 44.32 & 48.08 & 40.57 & 47.24 & 58.49 & 65.58 & 63.67 & 67.07 & 67.36 \\
    & (68.14\%) & 50K(full) & 71.86 & 73.08 & 71.33 & 72.38 & 70.85 & 71.77 & 73.36 & 73.84 & 73.27 \\
    \bottomrule
\end{tabular}}
\vspace{1mm}
\caption{
Ablation study of data size with top-1 classification accuracy (\%) of the student model on CIFAR-10 and CIFAR-100. 
}
\label{table:table4}
\end{center}
\vspace{-6mm}
\end{table*}
\setlength{\tabcolsep}{1.4pt}

\section*{B. More Results}

\subsection*{B.1. Complete Distillation Experiments}

We conduct different teacher-student model pairs for distillation experiments, and use ResNet32 / ResNet56 / VGG13 / ResNet110 / ResNet50 / ResNeXt101 as teacher models and use ResNet8 / ResNet32 / VGG11 / MobileNet / ResNet34 / ResNeXt50 as student models. 
Distillation performance is tested on various datasets, such as MNIST, CIFAR-10, CIFAR-100, Tiny ImageNet, and ImageNet-1K, as top-1 classification accuracy is exploited as an evaluation metric.
The experimental results are shown in Tab.~\ref{table:table1}, Tab.~\ref{table:table2} and Tab.~\ref{table:table3}.
For the training of teacher and student models, we adopt the same setting of hyperparameters, so as to verify the distillation effect of student models trained with different methods compared with the teacher model trained with vanilla supervised learning under the same conditions.

We also provide complete ablation results of different data sizes on CIFAR-10 and CIFAR-100, as shown in Tab.~\ref{table:table4}. 
We use an effective teacher-student pair of ResNet56 - MobileNet for experiments.
The results show that B2KD methods are generally more robust than traditional KD methods for small data sizes, and they can utilize the information in available samples maximumly to model compression in extreme cases.
In the comparison of all methods, MEKD achieves the best performance, which also validates the effectiveness and robustness of our proposed method.

\begin{figure}[th!]
\centering
\begin{minipage}{0.49\linewidth}
    \vspace{3pt}
    \centerline{\includegraphics[width=\textwidth]{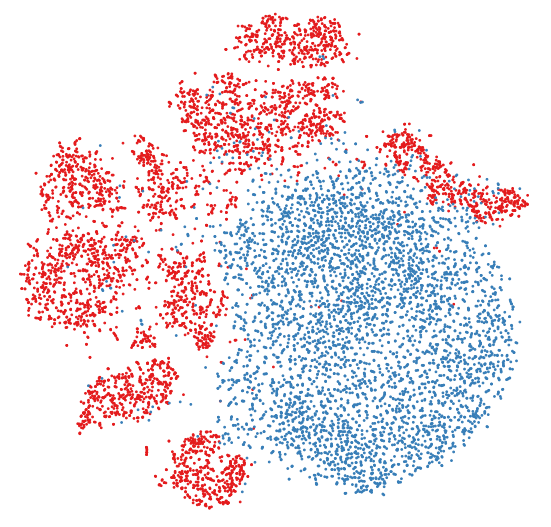}}
    \centerline{EPOCH 0}
\end{minipage}
\begin{minipage}{0.49\linewidth}
    \vspace{3pt}
    \centerline{\includegraphics[width=\textwidth]{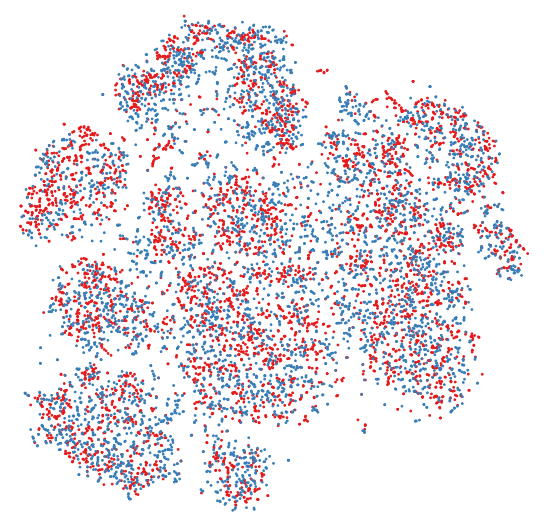}}
    \centerline{EPOCH 10}
\end{minipage}

\centering
\begin{minipage}{0.49\linewidth}
    \vspace{3pt}
    \centerline{\includegraphics[width=\textwidth]{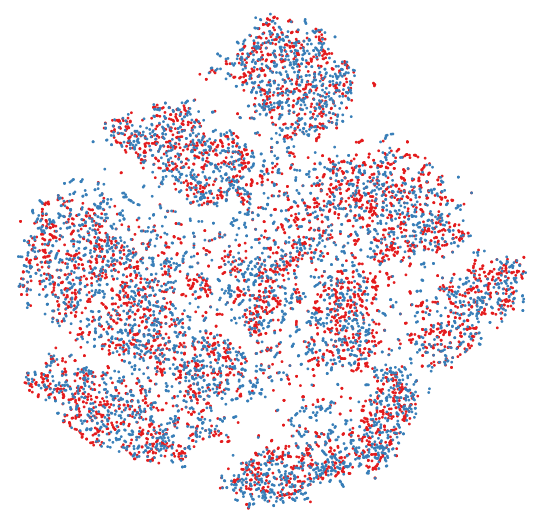}}
    \centerline{EPOCH 20}
\end{minipage}
\begin{minipage}{0.49\linewidth}
    \vspace{3pt}
    \centerline{\includegraphics[width=\textwidth]{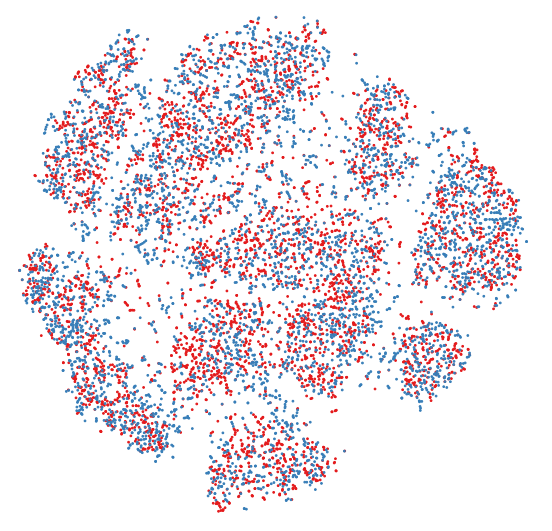}}
    \centerline{EPOCH 30}
\end{minipage}
\vspace{3mm}
\caption{t-SNE visualization of synthetic (blue) and genuine (red) images of MEKD with DCGAN on MNIST.}
\label{tsne1}
\vspace{-5mm}
\end{figure}

\begin{figure}[th!]
\centering
\begin{minipage}{0.49\linewidth}
    \vspace{3pt}
    \centerline{\includegraphics[width=\textwidth]{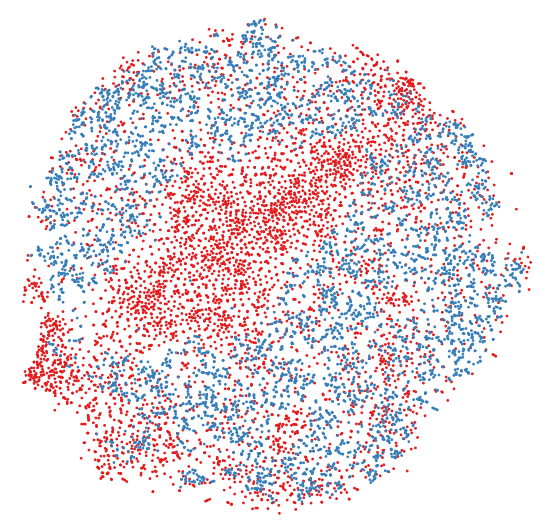}}
    \centerline{EPOCH 0}
\end{minipage}
\begin{minipage}{0.49\linewidth}
    \vspace{3pt}
    \centerline{\includegraphics[width=\textwidth]{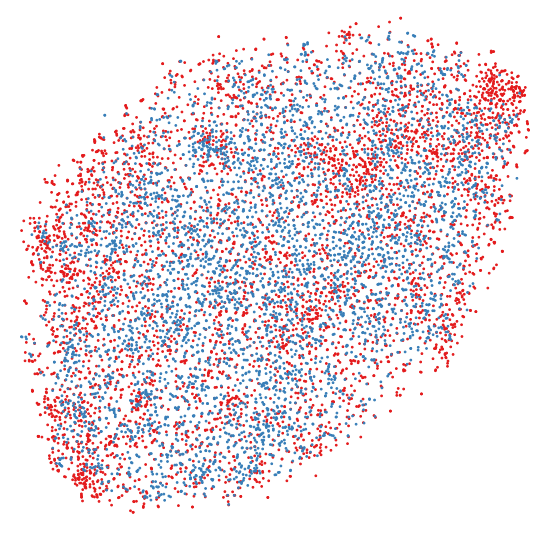}}
    \centerline{EPOCH 50}
\end{minipage}

\centering
\begin{minipage}{0.49\linewidth}
    \vspace{3pt}
    \centerline{\includegraphics[width=\textwidth]{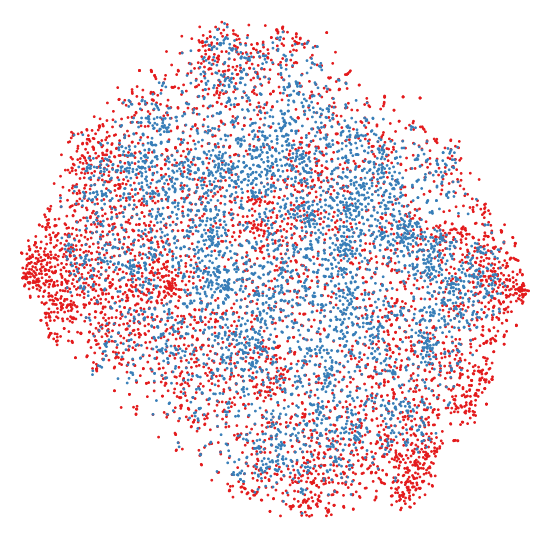}}
    \centerline{EPOCH 100}
\end{minipage}
\begin{minipage}{0.49\linewidth}
    \vspace{3pt}
    \centerline{\includegraphics[width=\textwidth]{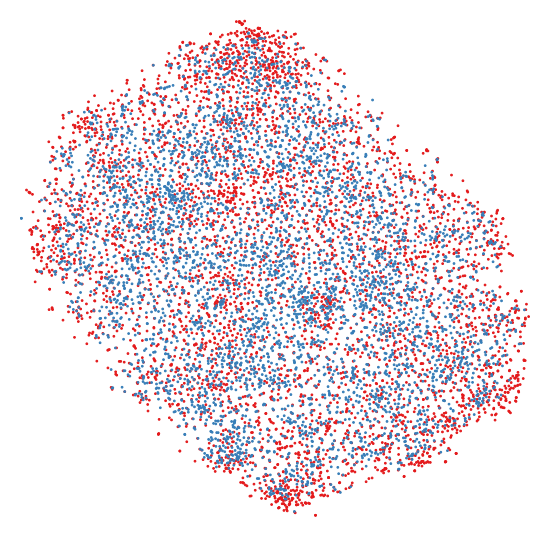}}
    \centerline{EPOCH 200}
\end{minipage}
\vspace{3mm}
\caption{t-SNE visualization of synthetic (blue) and genuine (red) images of MEKD with DCGAN on CIFAR-10.}
\label{tsne2}
\vspace{-6mm}
\end{figure}

In all experiments, teacher and student models are trained for $350$ epochs, except $12$ epochs for MNIST. We use Nesterov SGD with momentum $0.9$ and weight-decay $0.0005$ for training and use a mini-batch size of $128$ images on a single NVIDIA GeForce RTX 3090 GPU.
The initial learning rate is $0.1$, except $0.01$ for MNIST, and we conduct a multi-step learning rate schedule which decreases the learning rate by 0.1 at the $116^{th}$ and $233^{th}$ epoch for the training of models, except no learning rate schedule is used for MNIST.
For the training of student models, we follow the \emph{unsupervised} setting and only use the soft or hard responses of teacher models for distillation. 
Note that for all experiments, we conduct \emph{three} times experiments and report the mean accuracy. 

For the training of DCGAN, we follow the hyperparameters' settings of the work \cite{radford2015unsupervised}. 
DCGAN composes of a generator realized by transposed convolution layer and a discriminator realized by an ordinary convolution layer, which greatly reduces the number of network parameters and improves the image generation effect.
As an extension of our method, we believe that generative models of different architectures can also be used as emulators to learn the inverse mapping of the teacher function, by adding information maximization (IM) loss to alleviate the problem of mode collapse and achieve the purpose of deprivatization.
This will be our research work in the future.

\subsection*{B.2. Visualization Results}

We evaluate the training process of DCGAN in terms of whether the generated distribution is consistent with the real distribution, and visualize the synthetic and genuine images by t-SNE projection.
As shown in Fig.~\ref{tsne1} and Fig.~\ref{tsne2}, it can be observed that in the training process of DCGAN, the generated distribution is gradually closer to the real distribution. 
This verifies the effectiveness of using DCGAN as the emulator to learn the inverse mapping of the teacher function, and also proves that DCGAN can indeed alleviate the problem of mode collapse and generate images consistent with the distribution of real images. These synthetic images can not only effectively integrate various patterns in genuine images, but also serve as effective query samples to support the distillation of student models.

\end{document}

%% file: preamble.tex
\usepackage[dvipsnames]{xcolor}

\usepackage{times}
\usepackage{epsfig}
\usepackage{graphicx}
\usepackage{amsmath}
\usepackage{amssymb}
\usepackage{booktabs}
\usepackage{color}
\usepackage{float}
\usepackage{amsthm}
\usepackage{multirow}
\newtheorem{theorem}{Theorem}

\newtheorem{definition}{Definition}
\usepackage{amsmath,amssymb}
\usepackage{booktabs}
\usepackage{bbm}
\usepackage{algorithm}
\usepackage{algorithmic}
\usepackage{graphicx}

\definecolor{GRAY}{RGB}{128,128,128}

\usepackage[colorlinks,bookmarksopen,bookmarksnumbered,citecolor=blue, linkcolor=red, urlcolor=blue]{hyperref}

%% file: sec/0_abstract.tex
\begin{abstract}
Black-Box Knowledge Distillation (B2KD) is a formulated problem for cloud-to-edge model compression with invisible data and models hosted on the server. 
B2KD faces challenges such as limited Internet exchange and edge-cloud disparity of data distributions.
In this paper, we formalize a two-step workflow consisting of deprivatization and distillation, 
and theoretically provide a new optimization direction from logits to cell boundary different from direct logits alignment.
With its guidance, we propose a new method Mapping-Emulation KD (MEKD) that distills a black-box cumbersome model into a lightweight one.
Our method does not differentiate between treating soft or hard responses, and consists of:
1) deprivatization: emulating the inverse mapping of the teacher function with a generator, and 2) distillation: aligning low-dimensional logits of the teacher and student models by reducing the distance of high-dimensional image points.
For different teacher-student pairs, our method yields inspiring distillation performance on various benchmarks, and outperforms the previous state-of-the-art approaches.
\end{abstract}

%% file: sec/1_intro.tex
\section{Introduction}
\label{introduction}

Knowledge Distillation (KD) is a widely accepted approach to the problem of model compression and acceleration, which has received sustained attention from both the academic and industrial research communities \cite{gou2021knowledge,ozkara2021quped,stanton2021does,he2020group}.
The goal of KD is to extract knowledge from a cumbersome model or an ensemble of models, known as the teacher, and use it as supervision to guide the training of lightweight models, known as the student \cite{bergmann2020uninformed,pan2020spatio,aguilar2020knowledge}. 
In the application of KD, privacy protection has always been a very concerning issue for researchers and users, which not only refers to the privacy of user data but also includes the model copyright of cloud service providers.

\begin{figure}
  \centering 
  \includegraphics[scale=0.164]{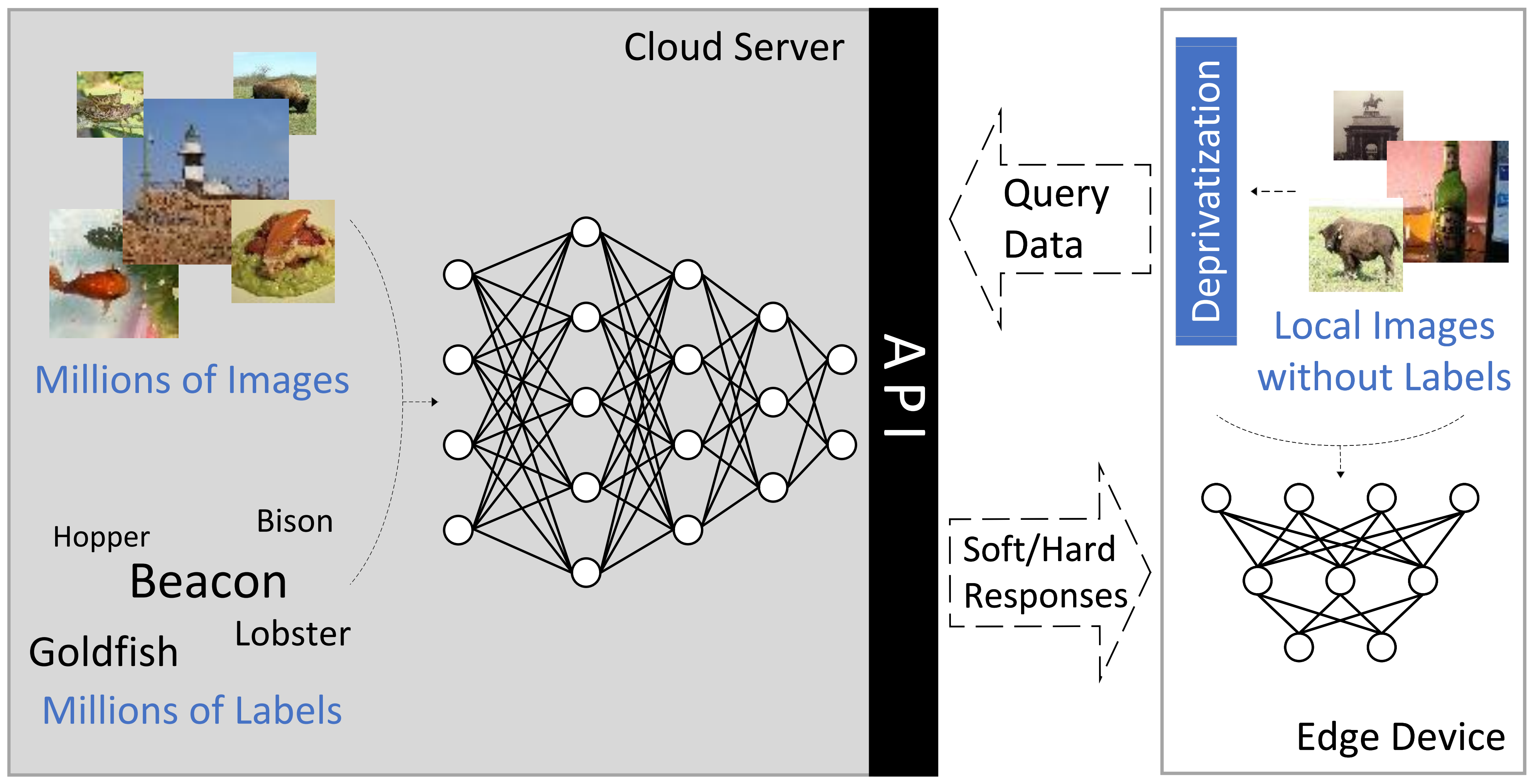}
  \vspace{-6mm}
  \caption{ 
  Schematic process of cloud-to-edge model compression. A cumbersome black-box model is deployed on a cloud server, trained with millions of samples and tags. The cloud server only provides APIs to receive query data and return inference responses of either soft or hard type. The edge device needs to distill a lightweight model using unlabeled local data.
  }
  \label{fig:intro}
  \vspace{-6mm}
\end{figure}

Black-Box Knowledge Distillation (B2KD) is a problem posed in the process of cloud-to-edge model compression \cite{orekondy2019knockoff,wang2020neural,wang2021zero}.
The cloud server hosts a teacher model whose internal structure and composition, connections between layers, model parameters, and gradients used for back-propagation are all invisible and unavailable to edge devices, as shown in Fig.~\ref{fig:intro}. 
Due to resource limitations, the edge device can only host a lightweight student model.
At the same time, low-quality and unlabeled local data cannot be used to train a reliable deep neural network. 
As a result, it must rely on sending query samples to the APIs of cloud servers for heavy inference \cite{tramer2016stealing}.

In practice, B2KD faces some key challenges. 
(a) Cloud servers and edge devices should maintain limited data exchange due to Internet latency and bandwidth constraints, as well as charges for the amount of queried data or API usage time.
(b) In some cases, for query samples, these APIs only provide indexes or semantic tags for the category with the highest probability (\emph{i.e.}, hard responses), rather than probability vectors for all possible classes (\emph{i.e.}, soft responses).
(c) Because users refuse to send sensitive data to cloud servers, the distribution gap between local and cloud data is difficult to measure, making the distilled student model inaccurate in the application.

Adversarial learning has been shown to be effective in generating pseudo samples, which is widely used in data augmentation and low-shot learning \cite{chen2019data,wang2021zero}. 
A well-trained generator can overcome the mode collapse problem and align real and synthetic data distribution.
In particular, we want to produce images relevant to training, whether or not they resemble real data \cite{micaelli2019zero}.
Meanwhile, images generated to obtain high responses from the teacher model combine different patterns with highly generalized features instead of sample-specific idiosyncrasies \cite{wang2020high}.
Therefore, using a well-trained generator to synthesize pseudo images can automatically filter out privacy-related high-frequency information, this process is called \textbf{deprivatization},

In this paper, we propose an approach to solve B2KD by \emph{mapping emulation}. 
Our motivation is in accordance with the fact that it can drive alignment between low-dimensional logits 
by reducing the distance between two generated images in the high-dimensional space. 
In addition, we argue that an image contains a lot of fine-grained information, which can be treated as another type of knowledge to provide different gradient directions for updating the parameters of student model, as shown in Fig.~\ref{fig:cell}. 
Combining image-level loss with coarse-grained logit-level loss can effectively improve the distillation effect.
According to the Kolmogorov theorem \cite{koppen2002training,braun2009constructive}, a sufficiently complex neural network is capable of representing an arbitrary multivariate continuous function from any dimension to another. 
Thus, a well-trained generator can not only emulate the inverse mapping of the teacher function (Thm.~\ref{theorem1}) but also help update the logits of a student to converge to the logits of a teacher (Thm.~\ref{theorem2}), with reasonable generalizability (Thm.~\ref{theorem3}).

In practice, we derive using a generative adversarial network (GAN) for deprivatization and exploit it as an inverse mapping of the teacher function.
The generator uses random variables as inputs that are sampled from a prior distribution with the same dimensionality as the logits. 
The well-trained generator is frozen and grafted behind the teacher and student model, whose output logits of the same examples are used as the inputs of the generator, as shown in Fig.~\ref{fig:pipeline}.
Experimental results show that MEKD can effectively protect the privacy of local data and models in the cloud, and it performs well under either soft or hard responses.
At the same time, MEKD has robust results in the case of limited query samples and out-of-domain data.

Overall, the contributions of this paper are:
{\bf 1)} We formalize the problem of B2KD and provide a two-step workflow of deprivatization and distillation.
{\bf 2)} We theoretically provide a new optimization direction from logits to cell boundary different from direct logits alignment.
{\bf 3)} We propose a new method of Mapping-Emulation Knowledge Distillation (MEKD).
The improved experimental performance has demonstrated the effectiveness of our approach.

%% file: sec/2_relatedwork.tex
\section{Related Work}
\label{relatedwork}

\textbf{Knowledge Distillation (KD)}. Hinton \emph{et al.} \cite{hinton2015distilling} propose an original teacher-student architecture that uses the logits of the teacher model as the knowledge. Since then, some KD methods regard knowledge as final responses to input samples \cite{ba2014deep,meng2019conditional,zhao2022decoupled}, some regard knowledge as features extracted from different layers of neural networks \cite{komodakis2017paying,kim2018paraphrasing,passban2021alp}, and some regard knowledge as relations between such layers \cite{yim2017gift,passalis2020heterogeneous,chen2020learning}. The purpose of defining different types of knowledge is to efficiently extract the underlying representation learned by the teacher model from the large-scale data. If we consider a network as a mapping function of input distribution to output, then different knowledge types help to approximate such a function.
Based on the type of knowledge transferred, KD can be divided into response-based, feature-based, and relation-based \cite{gou2021knowledge}. 
The first two aim to derive the student to mimic the responses of the output layer or the feature maps of the hidden layers of the teacher, and the last approach uses the relationships between the teacher's different layers to guide the training of the student model. 
Feature-based and relation-based methods \cite{komodakis2017paying,yim2017gift}, depending on the model utilized, may leak the information of structures and parameters through the intermediate layers' data. 
For example, we can reconstruct a ResNet \cite{he2016deep} based on the feature dimensions 
of different layers, and calculate each neuron's parameter using specific images and their responses in the feature maps.

\textbf{Black-Box Knowledge Distillation (B2KD)}.
Response-based KD methods \cite{hinton2015distilling,zhao2022decoupled,ba2014deep} have the natural property of hiding models. 
Hinton \emph{et al.} \cite{hinton2015distilling} use  Kullback-Leibler Divergence (KLD) between the softened logits of teacher and student models as the loss to align the output distribution, and Zhao \emph{et al.} \cite{zhao2022decoupled} decouple the KLD into two uncorrelated losses and combine them by weighted summation. 
These calculations do not take into account the details of the teacher model, which is exactly a black box.
The recently proposed approaches for B2KD also address the issue of hiding the teacher model deployed in the cloud server \cite{orekondy2019knockoff,wang2020neural,wang2021zero}.
Orekondy \emph{et al.} \cite{orekondy2019knockoff} use a reinforcement learning approach to improve query sample efficiency.
Wang \emph{et al}. \cite{wang2020neural} blend mixup and activate learning to augment the few unlabeled images and choose hard examples for distillation.
And Wang \cite{wang2021zero} proposes a decision-based black-box model and constructs the soft label for each training sample by computing its distances to the decision boundaries of the teacher model.
These existing approaches partially address the challenges of cloud-to-edge black-box model distillation, but none of them take into account the privacy leak of user data when sending original local images to the cloud.

\begin{figure*}
\vspace{-2mm}
  \centering 
  \includegraphics[scale=0.216]{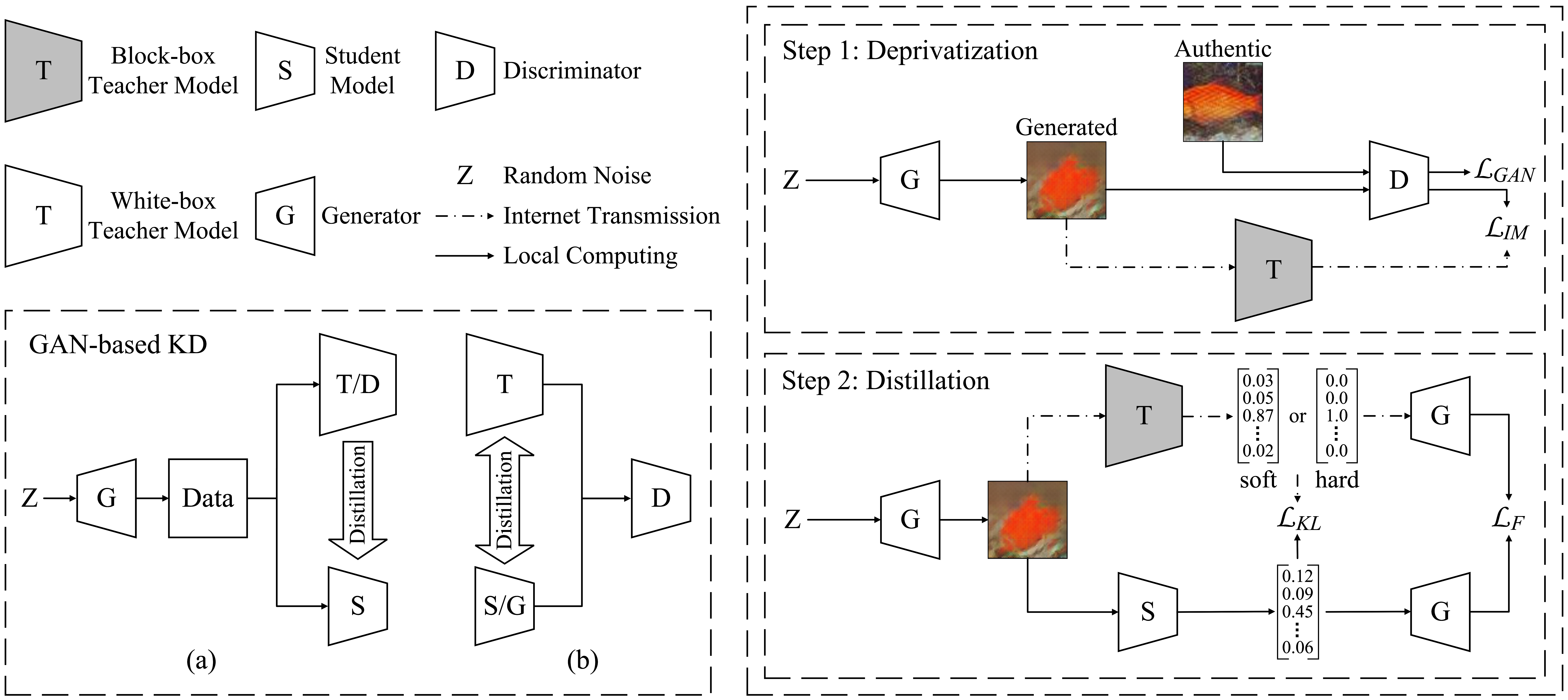}
  \vspace{-2mm}
  \caption{
  The overall framework of MEKD. Lower left: two architectures of GAN-based KD. Upper right: the process of deprivatization. GAN is used to synthetic high-response images to the teacher model within the distribution of data in edge devices. Lower right: the process of distillation with the frozen generator. The synthetic privacy-free images are query samples sent to the teacher model through the APIs of cloud servers. The student model is distilled by reducing the logit-level and image-level discrepancy.
  }
  \label{fig:pipeline}
  \vspace{-5.5mm}
\end{figure*}

\textbf{Generative Adversarial Networks (GANs)} 
have the capacity to handle sharp estimated density functions and generate realistic-looking images efficiently. A typical GAN \cite{goodfellow2014generative} comprises a discriminator distinguishing real images and generated images, and a generator synthesizing images to fool the discriminator. GANs are divided into architecture-variant and loss-variant. The former focuses on network architectures \cite{radford2015unsupervised,brock2018large} or latent space \cite{mirza2014conditional,chen2016infogan}, \emph{e.g.}, some specific architectures are proposed for specific tasks \cite{zhu2017unpaired,karras2019style}. The latter utilizes different loss types and regularization tools \cite{arjovsky2017wasserstein,gulrajani2017improved} to enable more stable learning.

\textbf{Adversarial Distillation (AD)} 
exploits adversarial architecture to help the teacher and student model have a better understanding of the real data distribution \cite{gou2021knowledge,xu2017training,wang2018adversarial,belagiannis2018adversarial,shen2019meal}. 
The methods of AD can be divided into two types according to the generator-discriminator architecture, as shown in Fig.~\ref{fig:pipeline}: 
(a) the generator is used to synthetic images to obey a real distribution, and these images are used to help distill models \cite{chen2019data,wang2018adversarial};
(b) the teacher and student models are regarded as generators and another discriminator is drafted behind them to judge whether the distribution of features or logits is consistent \cite{xu2017training,ye2020data}.
AD is also employed for low-shot knowledge distillation and received inspiring results \cite{chen2019data}.
Our method provides an alternative adversarial architecture, which utilizes a well-trained generator to guide the alignment between the outputs of models.

%% file: sec/3_method.tex
\section{Theory for Mapping-Emulation KD}
\label{3.2}

\begin{figure} 
\vspace{1mm}
  \centering 
  \includegraphics[scale=0.52]{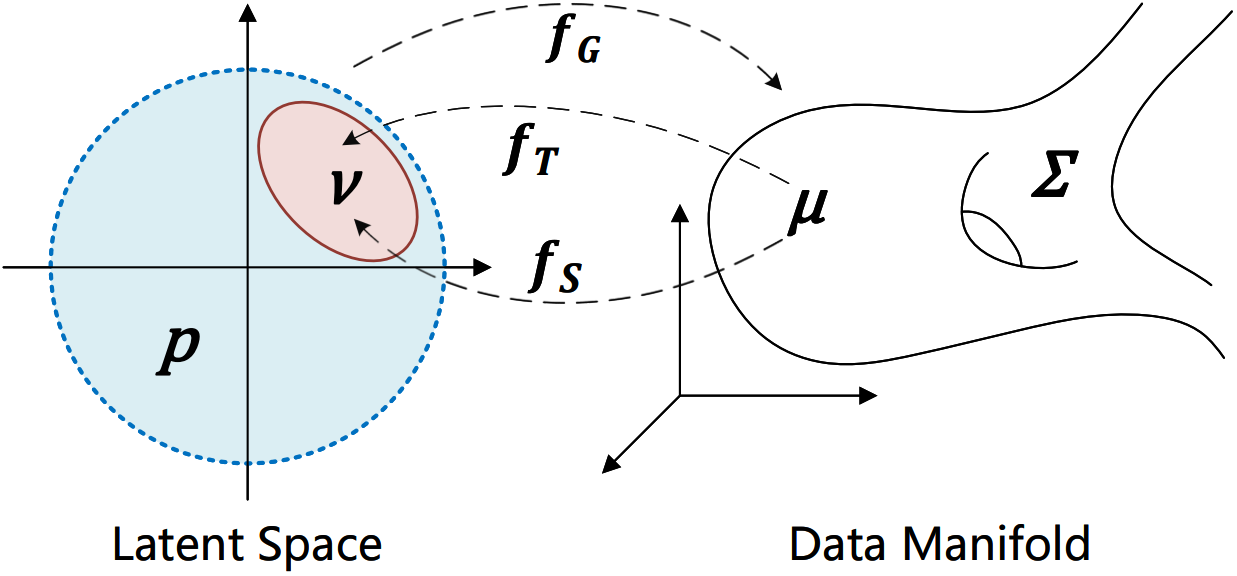}
  \vspace{-2mm}
  \caption{Mapping relationships of $f_S,f_T,f_G$. If $f_S$ and $f_T$ can map $\mu$ to the same distribution $\upsilon$, then $f_S=f_T$, and if $f_G$ can map the prior distribution $p$ to $\mu$, then $f_G=f_T^{-1}$.}
  \label{fig:mapping relationship}
  \vspace{-5mm}
\end{figure}

First, we propose two definitions. Def.~\ref{definition1} defines that two functions that map the same data distribution $\mu$ to the same latent distribution $\upsilon$ are equivalent. The ideal state of KD is to obtain a student function $f_S$ that is equivalent to the teacher function $f_T$. Def.~\ref{definition2} defines that the mapping function of a generator $G$, which can map a prior distribution $p$ to data manifold $\Sigma$ and guarantee that the generated image distribution $\mu'$ is the same as the real image distribution $\mu$, is considered to be the inverse mapping of the teacher function, \emph{i.e.} $f_G=f_T^{-1}$. And we call it a \emph{well-trained} generator. The mapping relationships are shown in Fig.~\ref{fig:mapping relationship}. 

\begin{definition}
\label{definition1} {\bf(Function Equivalence)}
Giving the student and teacher model $f_S$ and $f_T$, for a data distribution $\mu\in\mathcal{X}$ in image space which is mapped to $\mathbb{P}_S\in\mathcal{Y}$ and $\mathbb{P}_T\in\mathcal{Y}$ in latent space. If the Wasserstein distance between $\mathbb{P}_S$ and $\mathbb{P}_T$ equals zero,
\vspace{-2mm}
\begin{equation}
\vspace{-1mm}
    W(\mathbb{P}_S, \mathbb{P}_T)=\inf_{\gamma\in\Pi(\mathbb{P}_S, \mathbb{P}_T)}\mathbb{E}_{(y_S,y_T)\sim\gamma}\left[\ \|y_S-y_T\| \ \right] = 0,
\end{equation}
the student and teacher model are equivalent, \emph{i.e.}, $f_S=f_T$, where $\Pi(\mathbb{P}_S,\mathbb{P}_T)$ is the set of all joint distributions $\gamma(y_S,y_T)$ whose marginals are $\mathbb{P}_S$ and $\mathbb{P}_T$, respectively.
\end{definition}

\begin{definition}
\label{definition2} {\bf(Inverse Mapping)}
Giving a prior distribution $p\in\mathbb{R}^C$, for a data distribution $\mu\in\mathbb{R}^n$, if the Wasserstein distance between generated distribution $\mu'=(f_G)_{\#}p$ and $\mu$ equals zero,
\vspace{-1mm}
\begin{equation}
\vspace{-2mm}
    W(\mu', \mu)=\inf_{\gamma\in\Pi(\mu',\mu)}\mathbb{E}_{(x',x)\sim\gamma}[\ \|x'-x\|\ ]=0,
\end{equation}
then the generator $f_G:\mathbb{R}^C\rightarrow\mathbb{R}^n$ is the inverse mapping of the teacher function $f_T:\mathbb{R}^n\rightarrow\mathbb{R}^C$, denoted as $f_G=f_T^{-1}$, where $\Pi(\mu',\mu)$ is the set of all joint distributions $\gamma(x',x)$ whose marginals are respectively $\mu'$ and $\mu$.
\end{definition}

\begin{figure}
\vspace{-3mm}
\hspace{-1mm} 
  \centering 
  \includegraphics[scale=0.35]{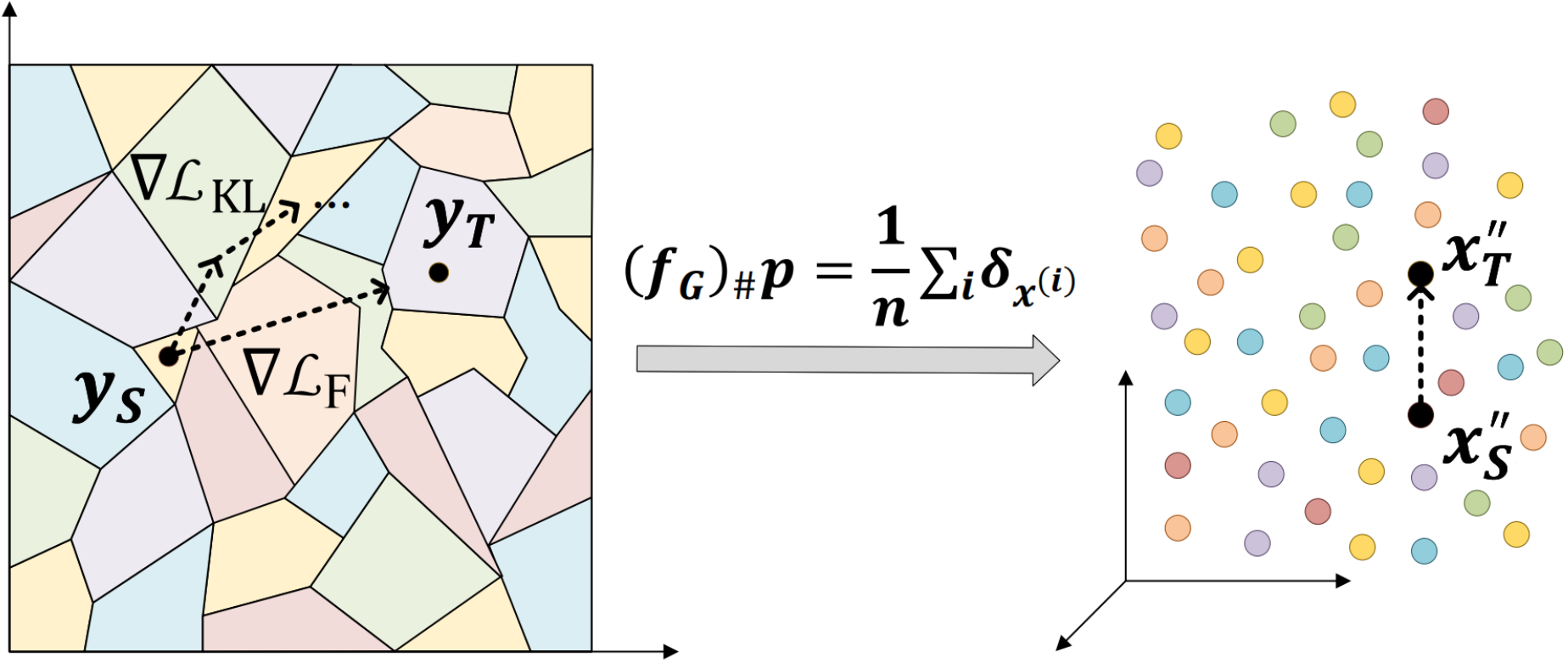}
  \vspace{-3mm}
  \caption{Cell $U_{\alpha}$ in the latent space is mapped via $f_G$ to an exact image $x^{(i)}$ of the same color. The move of point $x_S'$ to $x_T'$ causes the logits $y_S$ to align with $y_T$ from a direction different from $\mathcal{L}_{KL}$.}
  \label{fig:cell}
  \vspace{-4mm}
\end{figure}

Fixing a decoding map $f_G$ for a well-trained generator $G$, the latent space $\mathcal{Z}$ is partitioned as
\vspace{-1mm}
\begin{equation}
\vspace{-2mm}
    \mathcal{D}(f_G):\mathcal{Z}=\bigcup_{\alpha}U_{\alpha},
\end{equation}
\noindent where $\mathcal{D}(f_G)$ is called the decomposition induced by the decoding map $f_G$ \cite{lei2020geometric}, and $\{U_{\alpha}\}$ are called cells. 
As shown in Fig.~\ref{fig:cell}, $f_G$ maps a cell decomposition in the latent space $\mathcal{D}(f_G)$ to a cell decomposition in the image space $\frac{1}{n}\sum_{i}\delta_{x^{(i)}}$. Each cell $U_{\alpha}$ is mapped to a sample $\delta_{x^{(i)}}$ by the decoding map $f_G$ \cite{lei2019geometric}. In another word, $f_G$ pushes the prior distribution $p$ to the exact empirical distribution,
\vspace{-1mm}
\begin{equation}
\vspace{-1mm}
    ({f_G})_{\#}p=\frac{1}{n}\sum_{i}\delta_{x^{(i)}}.
\end{equation}

\begin{theorem}
\label{theorem1} {\bf(Empirical Approximation)}
For any $0<\epsilon<1/2$ and any integer $m>4$, let $g:\mathbb{R}^C\rightarrow\mathbb{R}^n$ be the mapping function of generator $G$ with $n\leq\frac{20\log m}{\epsilon^2}$. For two sets $V_S=\{y_S:y_S\in \mathbb{P}_S\}$ and $V_T=\{y_T:y_T\in \mathbb{P}_T\}$, both of which have $m$ points in $\mathbb{R}^C$, if the empirical Wasserstein distance between $g(V_S)$ and $g(V_T)$ equals zero,
\vspace{-2mm}
\begin{equation}
\vspace{-2mm}
    \hat{W}(g(V_S),g(V_T))=\frac{1}{m}\sum_{i=1}^m\|g(y_S^i)-g(y_T^i)\|=0,
\end{equation}
then $W(\mathbb{P}_S, \mathbb{P}_T)=0$.
\end{theorem}

Thm.~\ref{theorem1} (see Appendix for proof) provides a method to approximate the expected Wasserstein distance $W(\mathbb{P}_S,\mathbb{P}_T)$ using the empirical Wasserstein distance $\hat{W}(g(V_S),g(V_T))$. 
By reducing the distance between points $g(y_S^i)$ and $g(y_T^i)$ in high-dimensional space, an optimization direction $\nabla\mathcal{L}_F$ different from $\nabla\mathcal{L}_{KL}$ is produced for logits $y_S^i$ and $y_T^i$ in low-dimensional space. The gradient update causes $y_S^i$ to move towards the boundary of the cell in which $y_T^i$ resides, as shown in Fig.~\ref{fig:cell}.

\begin{theorem}
\label{theorem2} {\bf(Optimization Direction)}
Let $\mu\in\mathcal{X}$ be any distribution. $f_S, f_T, f_G$ are the mapping functions of the student, teacher, and generator, respectively. $f_S$ is parameterized by $\theta_S\in\Theta_S$. Then, when 
\vspace{-1mm}
\begin{equation}
\vspace{-1mm}
    \min_{\theta_S\in\Theta_S} \mathbb{E}_{x\sim \mu} \left[\|f_G\circ f_S(x), f_G\circ f_T(x)\|\right]\rightarrow 0,
\end{equation}
it holds that $f_S\rightarrow f_T$, and we have
\vspace{-1mm}
\begin{align}
\vspace{-1.5mm}
    &\nabla_{\theta_S}\mathbb{E}_{x\sim\mu}[f_S(x)]=\nabla_{\theta_S}W(\mathbb{P}_S, \mathbb{P}_T)\nonumber \\
    &\ \ \ \ \ \ \ \ =\mathbb{E}_{x\sim\mu}[\nabla_{\theta_S}\|f_G\circ f_S(x)-f_G\circ f_T(x)\|].
\end{align}
\end{theorem}

Thus, to achieve $f_S\rightarrow f_T$, it is sufficient to optimize $\mathbb{E}_{x\sim \mu} \left[\|f_G\circ f_S(x), f_G\circ f_T(x)\|\right]$ in the parameter space $\Theta_S$. 
The global gradient of parameter $\theta_S$ can be replaced by the gradient calculated on the empirical distance of high-dimensional image points, refer to Appendix for proof.

\begin{theorem}{\bf(Generalization Bound)}
\label{theorem3}
Let $H\subseteq \mathbb{R}^{\mathcal{X}\times\mathcal{Y}}$ be a hypothesis set for $C$-way classification task. For any $0<\epsilon<1/2$ and a sample $S$ of size $m>4$ drawn according to $\mu$, let $g:\mathbb{R}^C\rightarrow\mathbb{R}^n$ be a mapping function of generator $G$ with $n\leq\frac{20\log m}{\epsilon^2}$. Fix $\rho>0$, for any $1>\delta>0$, with probability at least $1-\delta$, the following holds for all $h\in H$,
\vspace{-2mm}
\begin{equation}
\vspace{-1mm}
    R(h)\le \hat{R}_{\rho}(h)+\frac{2C^2}{\rho(1-\epsilon)}\sqrt{\frac{r^2\Lambda^2}{m}}+\sqrt{\frac{\log\frac{1}{\delta}}{2m}}.
\end{equation}
For any $x\in\mathcal{X}$, the $\Lambda \geq 0$ and $(\sum_{y=1}^{C} \|h(x,y)\|^p)^{1/p}\leq \Lambda$ for any $p \geq 1$, and the $r>0$ for $K(x,x)\leq r^2$ where kernel $K:\mathcal{X}\times\mathcal{X}\rightarrow\mathbb{R}$ is positive definite symmetric.
\end{theorem}

Thm.~\ref{theorem3} (see Appendix for proof) gives the generalization bound of aligning low-dimensional logits by reducing the distance of high-dimensional image points, which guarantees generalizability to the unseen samples.

\section{Algorithm of Mapping-Emulation KD}
\label{3.1}

Hinton \emph{et al.} \cite{hinton2015distilling} propose a simple but effective KD method that uses the softened logits of the teacher model as a supervision to guide student training.
They use the Kullback-Leibler Divergence (KLD) to measure the discrepancy between the logits of the two models, where the student model is trained to minimize the gap in the hope of achieving the same output. The loss is defined as
\vspace{-1mm}
\begin{equation}
\vspace{-2mm}
\begin{aligned}
\label{equation:KLD}
    \mathcal{L}_{KL}&=\mathcal{KL}[p(c|\mathbf{x}_i;\theta_T)||p(c|\mathbf{x}_i;\theta_S)] \\ 
    &=\frac{1}{N}\sum_i^N\sum_c^C p(c|\mathbf{x}_i;\theta_T)\log\left[\frac{p(c|\mathbf{x}_i;\theta_T)}{p(c|\mathbf{x}_i;\theta_S)}\right],
\end{aligned}
\end{equation}
where $i$ is the sample index and $N$ is the number of samples.
Regardless of the method used, the essence of KD is to learn the mapping function of the teacher model from input to output, \emph{i.e.}, $f_T$.
However, it is hard to deduce the mapping function from the existing parameters of the teacher model. 
One can only guess the mapping process by using the responses to the input samples of different network layers or the relations between features and treat them as \emph{knowledge} to guide the training of the student model \cite{yim2017gift}. 
However, in the black-box KD problem, the internal responses or relations between layers of the teacher model are not available, which makes effective distillation more challenging.

\textbf{Deprivatization.}
For a $C$-way classification problem, we first train a GAN using the random noise variable $z$ sampled from the prior distribution $p$ in latent space $\mathcal{Y}$ as input. 
Note that the dimensionality of $z$ is the same as the output logits of the teacher model, \emph{i.e.} $|z|=C$. 
The generator $G$ uses noise $z$ to synthesize images, and the discriminator $D$ minimizes the Wasserstein distance between the generated $\mu'$ and the real distribution $\mu$. 
The synthetic \textit{privacy-free} images are simultaneously sent to the cloud server for inference responses, which can be soft (probability vectors for all possible classes) or hard (indexes or semantic tags for the category with the highest probability). 
We expect the synthetic images to match the high responses of the teacher model so that they can maximize the containment of patterns in real data. We adopt the information maximization (IM) loss \cite{hu2017learning,shi2012information}, which is formulated as
\vspace{-1mm}
\begin{equation}
\label{equation:lossofIM}
\vspace{-2mm}
    \mathcal{L}_{IM}=-\frac{1}{m}\sum_{i=1}^m \hat{y}^{(i)}_t \log\left(D\left(G\left(z^{(i)}\right)\right)\right),
\end{equation}
\noindent where $\hat{y}^{(i)}_t=\max_{c\in C} T\left(G\left(z^{(i)}\right)\right)$ for $0.0\le\hat{y}^{(i)}_t\le 1.0$ in soft responses and $\hat{y}^{(i)}_t=1.0$ in hard responses.

Suppose the discriminator is capable of completely blurring the discrepancy between synthetic and genuine images. 
In this case, the resulting generator represents a function from the latent space to the image space, defined as $f_G:\mathcal{Y}\rightarrow\mathcal{X}$, with an inverse mapping of the teacher function. 
Note that the generator and the discriminator are trained simultaneously: we adjust parameters for the generator to minimize $\log (1-D(G(z)))$ and adjust parameters for the discriminator to minimize $\log D(x)$. And their loss functions are
\vspace{-2mm}
\begin{equation}
\label{equation:lossofD}
\vspace{-1mm}
    \mathcal{L}_D=-\frac{1}{m}\sum_{i=1}^m\left[ \log D\left(x^{(i)}\right)+\log \left(1-D\left(G\left(z^{(i)}\right)\right)\right) \right],
\end{equation}
\begin{equation}
\label{equation:lossofG}
    \mathcal{L}_G=-\frac{1}{m}\sum_{i=1}^m\log \left(1-D\left(G\left(z^{(i)}\right)\right)\right).
\end{equation}
We introduce a trade-off hyperparameter $\alpha$ to balance $\mathcal{L}_{GAN}$ and $\mathcal{L}_{IM}$, and all the losses in the first step of deprivatization constitute
\vspace{-1mm}
\begin{equation}
\label{equation:lossofStep1}
\vspace{-1mm}
    \mathcal{L}_{Dp}=(\mathcal{L}_{G}+\mathcal{L}_D)+\alpha\mathcal{L}_{IM},
\end{equation}

\begin{algorithm}[b]
\caption{MEKD optimization algorithm.}
\label{alg:A}  
\hspace*{0.02in}{\bf Input:}
Pre-trained teacher $T(x;\theta_T)$ deployed in the cloud server, random initialized student $S(x;\theta_S)$ and local dataset $X$ hosted in the edge device.\\
\hspace*{0.02in}\vspace*{0.03in}{\bf Output:}
An optimized student $S(x;\theta_S)$ on dataset $X$.

\begin{algorithmic}[1]
\label{alg:mekd}
\STATE {\textcolor{GRAY}{$\triangleright$ Step 1: Deprivatization}}
\STATE {Initialize a generator $G(z;\theta_G)$ and a discriminator $D(x;\theta_D)$, and ensure the dimensionality of $z$ equals to the category count $C$.}
\STATE {\textbf{repeat}}
\STATE \hspace*{0.2in}{Sample a batch of noises $\mathcal{Z}$ from a prior distribu-}
\STATE \hspace*{0.2in}{tion $p$ and synthetic images $\mathcal{X}'=G(\mathcal{Z})$.}
\STATE \hspace*{0.2in}{The $\mathcal{X}'$ is sent to $T$ in cloud to get soft or hard}
\STATE \hspace*{0.2in}{inference responses $\hat{\mathcal{Y}}'_t=T(\mathcal{X}')$.}
\STATE \hspace*{0.2in}{Sample a batch of examples $\mathcal{X}$ from dataset $X$.}
\STATE \hspace*{0.2in}{Update the discriminator $D$ to distinguish $\mathcal{X}$}
\STATE \hspace*{0.2in}{and $\mathcal{X}'$ using $\mathcal{L}_D$ from Eqn.~\ref{equation:lossofD}.}
\STATE \hspace*{0.2in}{Update the generator $G$ to fool the discriminator $D$}
\STATE \hspace*{0.2in}{using $\mathcal{L}_G + \alpha\mathcal{L}_{IM}$ from Eqn.~\ref{equation:lossofIM} and Eqn.~\ref{equation:lossofG}.}
\STATE {\textbf{until} converge}
\STATE {\textcolor{GRAY}{$\triangleright$ Step 2: Distillation}}
\STATE {Initialize the student $S$ and freeze the generator $G$.}
\STATE {\textbf{repeat}}
\STATE \hspace*{0.2in}{Sample a batch of noises $\mathcal{Z}$ from a prior distribu-}
\STATE \hspace*{0.2in}{tion $p$ and synthetic images $\mathcal{X}'=G(\mathcal{Z})$.}
\STATE \hspace*{0.2in}{The $\mathcal{X}'$ is sent to $T$ in cloud to get soft or hard}
\STATE \hspace*{0.2in}{inference responses $\hat{\mathcal{Y}}'_t=T(\mathcal{X}')$.}
\STATE \hspace*{0.2in}{Update the student $S$ using $\mathcal{L}_{Dt}$ from Eqn.~\ref{equation:lossofStep2}.}
\STATE {\textbf{until} converge}
\end{algorithmic}
\end{algorithm}

\setlength{\tabcolsep}{4pt}
\renewcommand\arraystretch{1.1}
\begin{table*}[h!]
\begin{center}
\setlength{\tabcolsep}{1.2mm}{
\scalebox{.5}{}
\begin{tabular}{cc|cc|cc|cc|cc}
    \toprule
    Method  & Data Size & \multicolumn{2}{c|}{MNIST}  & \multicolumn{2}{c|}{CIFAR-10} & \multicolumn{2}{c|}{CIFAR-100} & \multicolumn{2}{c}{Tiny ImageNet}\\
    \hline
    \multirow{2}{*}{Teacher} & \multirow{2}{*}{50K$\sim$100K} & ResNet32 & VGG13  & ResNet56 & ResNet56 & VGG13 & ResNet56 & ResNet110 & ResNet110\\
    & & 99.50 & 99.52 & 94.15 & 94.15 & 74.68 & 72.06  & 60.71 & 60.71\\
    \multirow{2}{*}{Student} & \multirow{2}{*}{50K$\sim$100K} & ResNet8 & VGG11 & ResNet8 & VGG11 & VGG11 & VGG11 & ResNet32 & MobileNet\\ 
    & & 99.24 & 99.41 & 87.74 & 91.81 & 69.12 & 69.12 & 55.47 & 56.07  \\
	\hline
	 KD \cite{hinton2015distilling} & 50K$\sim$100K & 99.33 & 99.44 & 86.58 & 82.25 & 70.88 & 67.97 & 54.14 & 57.85 \\
	 ML \cite{ba2014deep} & 50K$\sim$100K & 99.49 & 99.40 & 87.89 & 91.91 & 67.78 & 70.18 & 56.56 & 60.07 \\
	 AL \cite{wang2018adversarial} & 50K$\sim$100K & 99.37 & 99.26 & 87.25 & 91.97 & 69.92 & 71.13 & 46.02 & 51.29 \\
     DKD \cite{zhao2022decoupled} & 50K$\sim$100K & 99.33 & 99.43 & 86.61 & 92.42 & 67.32 & 70.10 & 55.99 & 59.43 \\
     DAFL \cite{chen2019data} & 0K & 96.42 & 97.00 & 60.67 & 66.03 & 43.78 & 48.32 & 38.44 & 40.93 \\
     \hline
     KN \cite{orekondy2019knockoff} & 10K & 98.61 & 98.81 & 80.62 & 82.41 & 57.83 & 55.64 & 48.92 & 50.22 \\
     AM \cite{wang2020neural} & 10K & 99.33 & \textbf{99.47} & 74.89 & 74.26 & 62.17 & 63.20 & 47.72 & 51.54 \\
     DB3KD \cite{wang2021zero} & 10K & 98.94 & 99.16 & 78.47 & 85.84 & 63.48 & 62.76 & 47.95 & 50.49 \\
	 MEKD (soft)  & 10K & \textbf{99.40} & 99.43 & \textbf{85.36} & \textbf{87.27} & \textbf{64.76} & 64.83 & \textbf{50.87} & \textbf{54.93} \\
    MEKD (hard) & 10K & 99.40 & 99.45 & 84.45 & 87.25 & 64.72 & \textbf{65.32} & 49.89 & 54.71 \\
    \bottomrule
\end{tabular}}
\vspace{-2mm}
\caption{Top-1 classification accuracy (\%) of the student model on MNIST, CIFAR-10, CIFAR-100 and Tiny ImageNet.}
\label{table:main}
\end{center}
\vspace{-9mm}
\end{table*}
\setlength{\tabcolsep}{1.4pt}

\textbf{Distillation.}
The well-trained generator $G$ contains the knowledge that the teacher uses to make inferences. 
It is equivalent to a teacher assistant transferring the teacher's knowledge to the student. Fig.~\ref{fig:pipeline} illustrates the architecture of MEKD. 
We freeze the generator and graft it behind the teacher and student model in the same way, using the softened logits of both models as the generator input. 
A batch of synthetic images $X'=\{{x'}^{(i)}=f_G(z^{(i)})\}_{i=1}^m$ is fed into the embedded network to output high-dimensional points in the same image space, simultaneously. 
The distance between the output high-dimensional points from the logits of the teacher model $X_T''=f_G\circ f_T(X')$ and the others from the student $X_S''=f_G\circ f_S(X')$ are measured by the distance measurement formula $\mathcal{L}_{F}=\mathbbm{d}(X_S'', X_T'')$. 
We minimize the distance $\mathcal{L}_{F}$ to drive the student model to mimic the output logits of the teacher model, and use $\mathcal{L}_1$-norm ($F=1$) of $X_S''$ and $X_T''$ as the loss function to distill the student,
\vspace{-1mm}
\begin{align}
\label{equation:lossofStep2}
    \mathcal{L}_{Dt}&=\frac{1}{m}\sum_{i=1}^m\  \left\| G\left(S\left({x'}^{(i)}\right)/\tau\right) - G\left(T\left({x'}^{(i)}\right)/\tau\right) \right\|_{F} \nonumber \\
    &\ \ \ \ \ +\beta\frac{1}{m}\sum_{i=1}^m\ T\left({x'}^{(i)}\right) \log \left( \frac{T\left({x'}^{(i)}\right)}{S\left({x'}^{(i)}\right)}\right),
\end{align}
\noindent where query sample ${x'}^{(i)}=G(z^{(i)})$ is generated from noise $z^{(i)}$ and temperature $\tau$ is used to soften the output logits.
Through the experiments, we found that $\mathcal{L}_2$-norm has a similar effect with $\mathcal{L}_1$-norm, refer to Tab.~\ref{ablationofnorm}.

We also add logit-level knowledge (Eqn.~\ref{equation:KLD}) to induce distillation and use a hyperparameter $\beta$ to balance these two losses. 
Unlike most KD methods, we do not use cross-entropy loss with ground-truth labels, due to its unavailability in edge devices. An algorithm is summarized in Alg.~\ref{alg:mekd}.

%% file: sec/4_experiment.tex
\section{Experiments}
\label{experiments}

\vspace{-0.5mm}
\subsection{Experiment Setup}
\vspace{-0.5mm}

In this section, we compare our method with response-based KD and black-box KD methods in an \emph{unsupervised} environment. 
Experimental results show that when the cross-entropy loss based on ground-truth labels is removed, the distillation performance of these methods decreases.

\textbf{Datasets Setup.} 
We conduct experiments on MNIST \cite{lecun1998gradient}, CIFAR \cite{krizhevsky2009learning}, Tiny ImageNet \cite{deng2009imagenet}, and ImageNet-1K \cite{deng2009imagenet}, all of which are widely used for image classification. 
While training B2KD methods, we randomly select $10K$ images ($100K$ for ImageNet-1K) from the training set, and all images in the test set (val set for ImageNet) are used as the benchmark to calculate accuracy.
For other approaches, except DAFL \cite{chen2019data} based on zero-shot learning, we use the whole training set.
We mainly use top-1 classification accuracy as an evaluation metric to assess the distillation effect.
To make a fair and intuitive comparison, we follow the same setup as previous B2KD methods in our \emph{main experiments}. 
However, we find that the original settings in the B2KD experiments do not represent the challenges raised in practical applications, so we add \emph{extended experiments} in Sec.~\ref{extened} to illustrate the practicability of our proposed method.

\textbf{Implementation.} See also the project page\footnote{\url{https://github.com/HAIV-Lab/MEKD}}.
We use ResNet \cite{he2016deep}, VGG \cite{simonyan2014very} and MobileNet \cite{howard2017mobilenets} as the backbone, and adopt standard data augmentation techniques (random crop and horizontal flip) and an SGD optimizer in all experiments. 
We consistently train the teacher and student model for $350$ epochs, except for $12$ epochs for MNIST, and we adopt a multi-step LR scheduler following the paper \cite{kim2018paraphrasing}. 
After training the teacher, we train a DCGAN  \cite{radford2015unsupervised} with Gaussian noise in the same dimension as the category counts. 
The output logits of teacher or student for samples in the same class follow a Gaussian distribution, and the logits center is the mean of the Gaussian.
Since the conversion between different Gaussian distributions is a linear process, using Gaussian as the prior distribution $p$ provides a smooth dual space for the student's logits update.

\textbf{Competing Methods.} 
In order to verify the effectiveness of our method, we compare several methods of response-based KD and black-box KD. 
We select KD \cite{hinton2015distilling} proposed by Hinton \emph{et al.} and ML \cite{ba2014deep} proposed by Ba and Caruana as the baselines, and we also compare the recently published DKD \cite{zhao2022decoupled} based on decoupled KLD.
For the two GAN-based KD frameworks summarized in Sec. \ref{relatedwork}, we choose AL \cite{wang2018adversarial} and DAFL \cite{chen2019data} as comparison methods.
Meanwhile, we compete with some black-box KD methods such as KN \cite{orekondy2019knockoff}, AM \cite{wang2020neural} and DB3KD \cite{wang2021zero}.
Of these methods, DB3KD and MEKD(hard) only utilize hard responses, while the other methods are based on soft responses.

\renewcommand\arraystretch{1.0}
\begin{table}[b]
\vspace{-5mm}
\begin{center}
\setlength{\tabcolsep}{1.4mm}{
\scalebox{.5}{}
\begin{tabular}{c|c|c|c|c|c}
    \toprule
    T - S   & KD & AL & AM & DB3KD & MEKD \\
    Pairs & (soft) & (soft) & (soft) & (hard) & (soft) \\
    \midrule
    RN50 - RN34  & 52.08 & 53.50 & 56.92 & 58.61 & \textbf{59.89} \\
    \midrule
    RX101 - RX50 & 54.90 & 50.88 & 55.64 & 59.90 & \textbf{61.21} \\
    \bottomrule
\end{tabular}}
\vspace{-2.5mm}
\caption{Top-1 classification accuracy (\%) of the student model on ImageNet-1K with data size $100K$. We use the pre-trained ResNet50 (76.13\%) and ResNeXt101 (79.32\%) as teachers. }
\label{imagenet}
\end{center}
\vspace{-5mm}
\end{table}

\vspace{-0.5mm}
\subsection{Performance Evaluation}
\vspace{-0.5mm}

On MNIST, CIFAR, and Tiny ImageNet, we use ResNet32/56/110 and VGG13 as the teacher model and use ResNet8/32, VGG11, and MobileNet as the student model. We compare the top-1 classification accuracy (ACC) of different teacher-student pairs, the results are shown in Tab.~\ref{table:main}.

On relatively easy tasks, such as MNIST and CIFAR-10, our proposed method has a small gap compared to response-baed KD methods that use the full training set. This makes sense in the applications of cloud-to-edge model compression because edge devices do not have a lot of capacity to store more than ten thousand pieces of data.



CIFAR-100 and Tiny ImageNet are more challenging.
These tasks contain far more patterns than MNIST and CIFAR-10, and data distributions are so complex that it is difficult for a generator to capture all the patterns.
However, as long as the mode collapse problem can be mitigated, it is possible to synthesize complex samples beneficial to distillation, so we exploit DCGAN \cite{radford2015unsupervised} as our generator.
DCGAN has a more stable training process and is more suitable for generating RGB images than a fully-connected GAN \cite{radford2015unsupervised}.
Experimental results show that MEKD can obtain an accuracy improvement of $5\%\sim10\%$ compared to other B2KD methods, and the accuracy of MEKD with soft or hard responses is similar, with a difference of less than $1\%$.

We also conduct experiments on large-scale datasets and sophisticated networks. 
On ImageNet-1K, we use two teacher-student (T-S) pairs of ResNet50 (RN50) - ResNet34 (RN34) and ResNeXt101 (RX101) - ResNeXt50 (RX50). 
All methods are trained using a subset of $100K$ samples.
The experimental results are shown in Tab.~\ref{imagenet}.

Uniformly, we set the number of query samples to $50K$ on CIFAR and MNIST, $300K$ on ImageNet, and discuss the performance impact of limited query samples in Sec.~\ref{limitedquery}.

\renewcommand\arraystretch{1.0}
\begin{table}[b!]
\vspace{-5mm}
\begin{center}
\setlength{\tabcolsep}{2.5mm}{
\scalebox{.5}{}
\begin{tabular}{c|c|c|c|c}
    \toprule
    Data Size & 0.1K & 1K & 10K & 50K (full) \\
    \midrule
    KD \cite{hinton2015distilling} & 16.74 & 31.25 & 70.90 & 90.43 \\
    AL \cite{wang2018adversarial} & 12.97 & 32.05 & 68.61 & 90.54 \\
    AM \cite{wang2020neural} & 48.31 & 62.05 & 73.65 & 86.33 \\
    DB3KD \cite{wang2021zero} & 43.05 & 64.28 & 81.67 & 92.46 \\
    MEKD (soft) & 49.04 & 69.84 & 86.85 & 93.48 \\
    MEKD (hard) & 47.12 & 68.66 & 86.53 & 93.09 \\
    \bottomrule
\end{tabular}}
\vspace{-2.5mm}
\caption{Ablation study of data size on CIFAR-10. We use the T-S pair of ResNet56 - MobileNet, and the full training set is $50K$.}
\label{datasize}
\end{center}
\vspace{-3.5mm}
\end{table}

\subsection{Ablation Study}

We choose an effective T-S pair \cite{mirzadeh2020improved} of ResNet56 - MobileNet for ablation studies unless otherwise stated.

\textbf{Ablation Study of Data Size.}
We explore the performance with different data sizes, the results are shown in Tab.~\ref{datasize}.
In general, B2KD methods have higher robustness to small data sizes than traditional KD methods, and in which MEKD achieves the highest distillation performance.

\begin{figure}[t]
\begin{minipage}{0.32\linewidth}
    \centerline{\includegraphics[width=\textwidth]{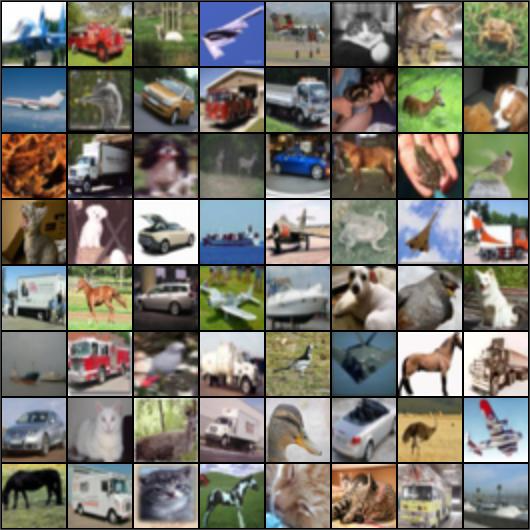}}
    \vspace{-1mm}
    \centerline{(a)}
\end{minipage}
\begin{minipage}{0.32\linewidth}
    \centerline{\includegraphics[width=\textwidth]{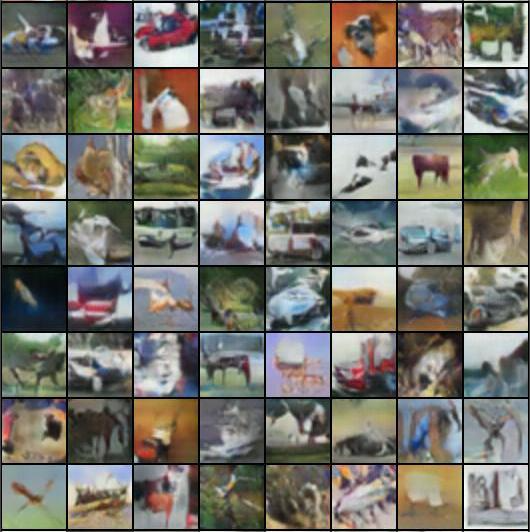}}
    \vspace{-1mm}
    \centerline{(b)}
\end{minipage}
\begin{minipage}{0.32\linewidth}
    \centerline{\includegraphics[width=\textwidth]{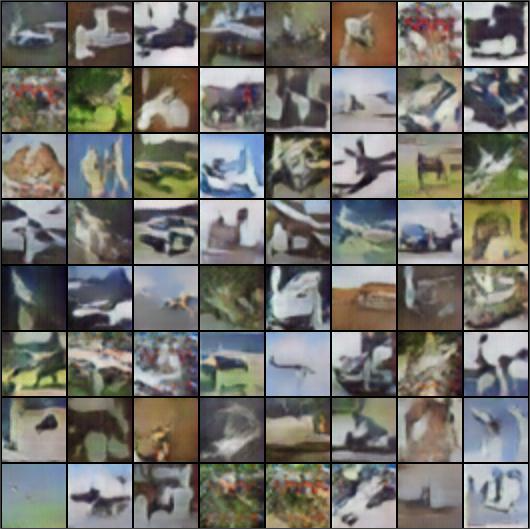}}
    \vspace{-1mm}
    \centerline{(c)}
\end{minipage}
\vspace{-3mm}
\caption{Real images of CIFAR-10 (a) and synthetic images using MEKD with $\mathcal{L}_{IM}$ (b) and without $\mathcal{L}_{IM}$ (c).}
\label{gen-imgs}
\vspace{-6.5mm}
\end{figure}

\textbf{Ablation Study of Deprivatization.}
The $\alpha$ is a hyperparameter to balance $\mathcal{L}_{GAN}$ and $\mathcal{L}_{IM}$. 
The $\mathcal{L}_{IM}$ is used to maximize the responses of the teacher to the generated samples.
Therefore, the training of the generator with or without $\mathcal{L}_{IM}$ will affect the quality of synthetic images. 
Fig.~\ref{gen-imgs} (a) shows real images of CIFAR-10.
Fig.~\ref{gen-imgs} (b) shows synthetic images with $\alpha=0.5$ and Fig.~\ref{gen-imgs} (c) shows synthetic images without $\mathcal{L}_{IM}$ (\emph{i.e.} $\alpha=0$), both using the same noise vectors.
The teacher of ResNet56 responds from $0.72\sim0.96$ to the synthetic image in Fig.~\ref{gen-imgs} (b) and from $0.41\sim0.87$ to the one in Fig.~\ref{gen-imgs} (c).
The effect of $\alpha$ is also reported in Tab.~\ref{ablationoftwo}, which reflects that the utilization of $\mathcal{L}_{IM}$ can improve the performance of model distillation.

\begin{figure}[b!]
\vspace{-6.5mm}
\centering
\hspace{-2mm}
\begin{minipage}{0.5\linewidth}
    \vspace{3pt}
    \centerline{\includegraphics[width=\textwidth]{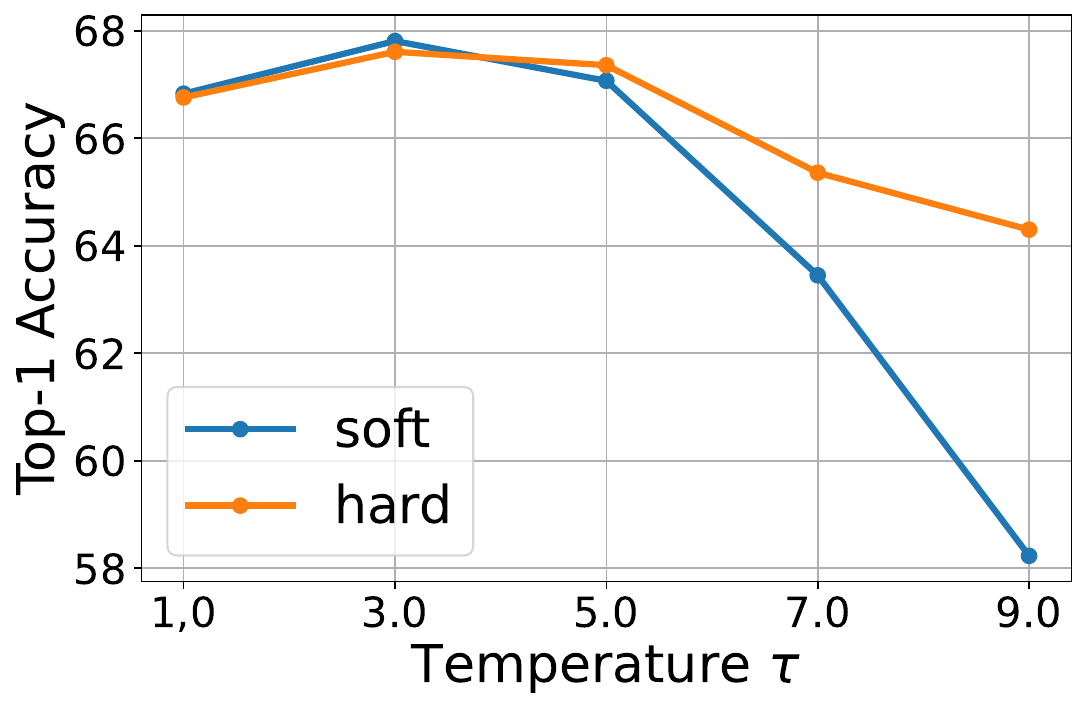}}
    \vspace{-2mm}
    \centerline{\hspace{5mm}(a)}
\end{minipage}
\begin{minipage}{0.5\linewidth}
    \vspace{3pt}
    \centerline{\includegraphics[width=\textwidth]{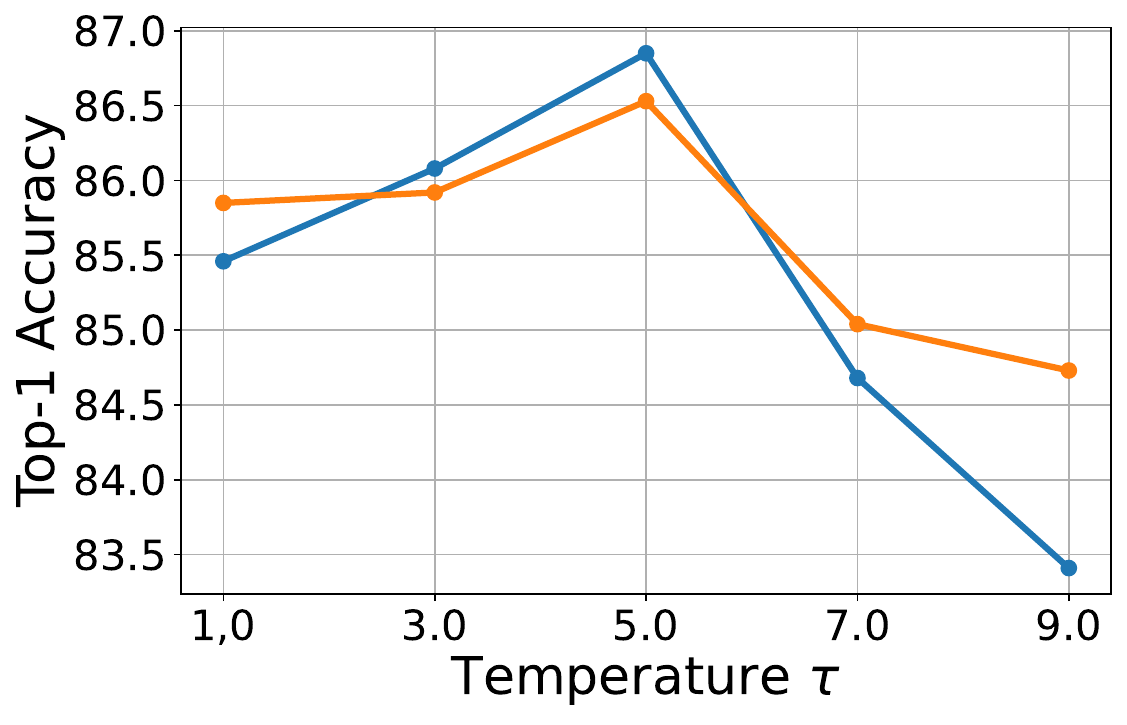}}
    \vspace{-2mm}
    \centerline{\hspace{6.5mm}(b)}
\end{minipage}
\vspace{-3.5mm}
\caption{Ablation study of temperature $\tau$ on CIFAR-100 (a) and CIFAR-10 (b).
We use the T-S pair of ResNet56 - MobileNet.
}
\label{ablationoftau}
\end{figure}

\renewcommand\arraystretch{1.0}
\begin{table}[b]
\vspace{-2mm}
\begin{center}
\setlength{\tabcolsep}{3mm}{
\centering
\begin{minipage}{0.23\textwidth}
\scalebox{.5}{}
\centering
\begin{tabular}{c|cc}
    \toprule
    \multirow{2}{*}{$\alpha$} & \multicolumn{2}{c}{Response Type} \\
     & Soft & Hard \\
    \midrule
    0.0 & 60.87 & 61.31 \\
    0.1 & 66.06 & 66.86 \\
    0.5 & 67.07 & 67.36 \\
    1.0 & 67.01 & 67.11 \\
    \bottomrule
\end{tabular}
\end{minipage}
\noindent
\centering
\begin{minipage}{0.23\textwidth}
\scalebox{.5}{}
\centering
\begin{tabular}{c|cc}
    \toprule
    \multirow{2}{*}{$\beta$} & \multicolumn{2}{c}{Response Type} \\
    & Soft & Hard \\
    \midrule
    0.0 & 56.28 & 56.23 \\
    0.1 & 64.13 & 65.60 \\
    0.5 & 66.79 & 67.01 \\
    1.0 & 67.07 & 67.36 \\
    \bottomrule
\end{tabular}
\end{minipage}
}
\end{center}
\vspace{-5.5mm}
\caption{
Ablation study of hyperparamete $\alpha$ and $\beta$ on CIFAR-100.
We use the T-S pair of ResNet56 - MobileNet.
}
\label{ablationoftwo}
\end{table}

\textbf{Ablation Study of Distillation.}
In Eqn.~\ref{equation:lossofStep2}, the $\beta$ is a trade-off hyperparameter to balance $\mathcal{L}_F$ and $\mathcal{L}_{KL}$, which provide different gradient directions for $\theta_S$.
As shown in Tab.~\ref{ablationoftwo}, the distillation performance can be improved by introducing KLD as an additional loss function.

The temperature $\tau$ is another important hyperparameter for MEKD since it softens the output logits of both the teacher and student models.
The results are shown in Fig.~\ref{ablationoftau}.
Its validity comes from the fact that softened logits can increase the probability of being sampled in a standard normal distribution.
Since GANs use a standard Gaussian distribution as input, samples generated from out-of-distribution noises with low-sampling probability are usually fuzzy and incorporate few patterns \cite{schlegl2017unsupervised}, which are meaningless for distillation.
Meanwhile, a high value of $\tau$ reduces the discrepancy between softened logits, and $\mathcal{L}_F=0$ when they locate in the same cell. It reduces the performance of distillation, especially for challenging tasks, such as ImageNet.

\textbf{Ablation Study of Different $\mathcal{L}_F$.}
In Eqn.~\ref{equation:lossofStep2}, we use $\mathcal{L}_F$ to calculate the distance between generated samples $X_S''$ and $X_T''$. 
From the analysis of experimental results, as shown in Tab.~\ref{ablationofnorm}, we argue that the effect on distillation is similar whether $F$ equals $1$ or $2$. 
The reason is that $\mathcal{L}_F$ is used to measure the distance between logits of the student and the boundary of cells, in which logits of the teacher reside, and different $\mathcal{L}_F$ represent similar gradient directions.


\renewcommand\arraystretch{1.0}
\begin{table}[t!]
\begin{center}
\setlength{\tabcolsep}{1.4mm}{
\scalebox{.5}{}
\begin{tabular}{c|c|c|c}
    \toprule
    Dataset & Method & Model & ACC ($\mathcal{L}_1$/$\mathcal{L}_2$)\\
    \midrule
    \multirow{2}{*}{CIFAR-10} 
    & MEKD (soft) & MobileNet & 86.85 / 86.63 \\
    & MEKD (hard) & MobileNet & 86.53 / 86.88 \\
    \midrule
    \multirow{2}{*}{CIFAR-100} 
    & MEKD (soft) & MobileNet & 67.07 / 66.95 \\
    & MEKD (hard) & MobileNet & 67.36 / 66.94 \\
    \bottomrule
\end{tabular}}
\vspace{-2.5mm}
\caption{
Ablation study of different $\mathcal{L}_F$.
We use ResNet56 as the teacher model. ACC: top-1 classification accuracy (\%).
}
\label{ablationofnorm}
\end{center}
\vspace{-8mm}
\end{table}

\begin{figure}[t]
\centering
\hspace{-2mm}
\begin{minipage}{0.5\linewidth}
    \vspace{3pt}
    \centerline{\includegraphics[width=\textwidth]{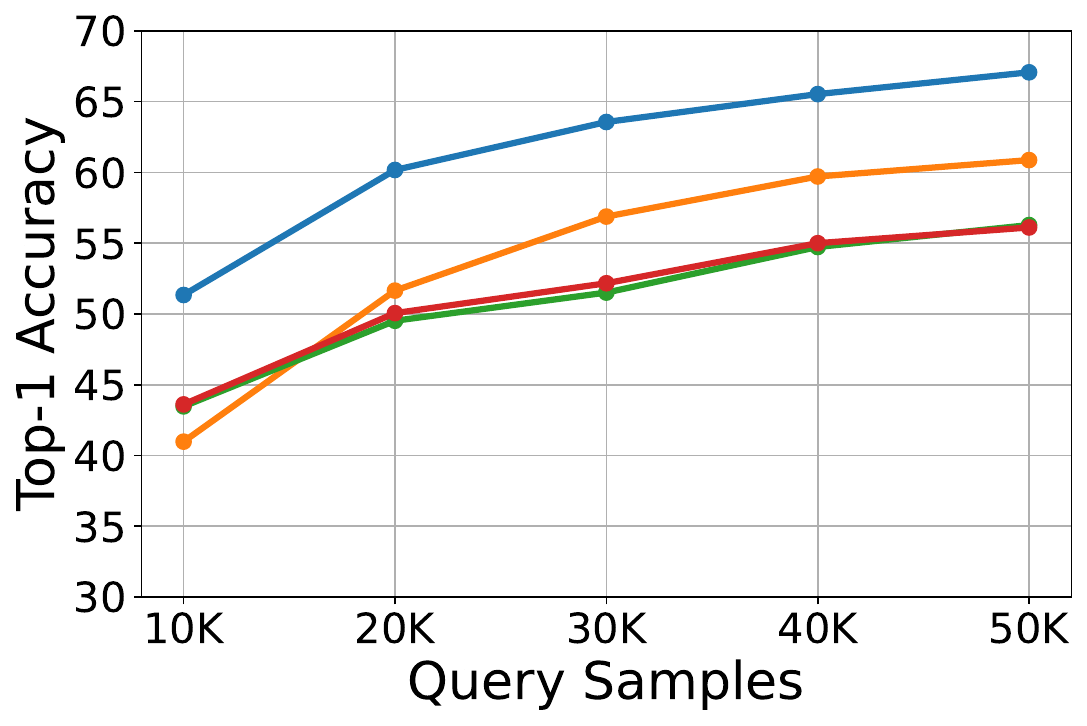}}
    \vspace{-1.5mm}
    \centerline{\hspace{5mm}(a)}
\end{minipage}
\begin{minipage}{0.5\linewidth}
    \vspace{3pt}
    \centerline{\includegraphics[width=\textwidth]{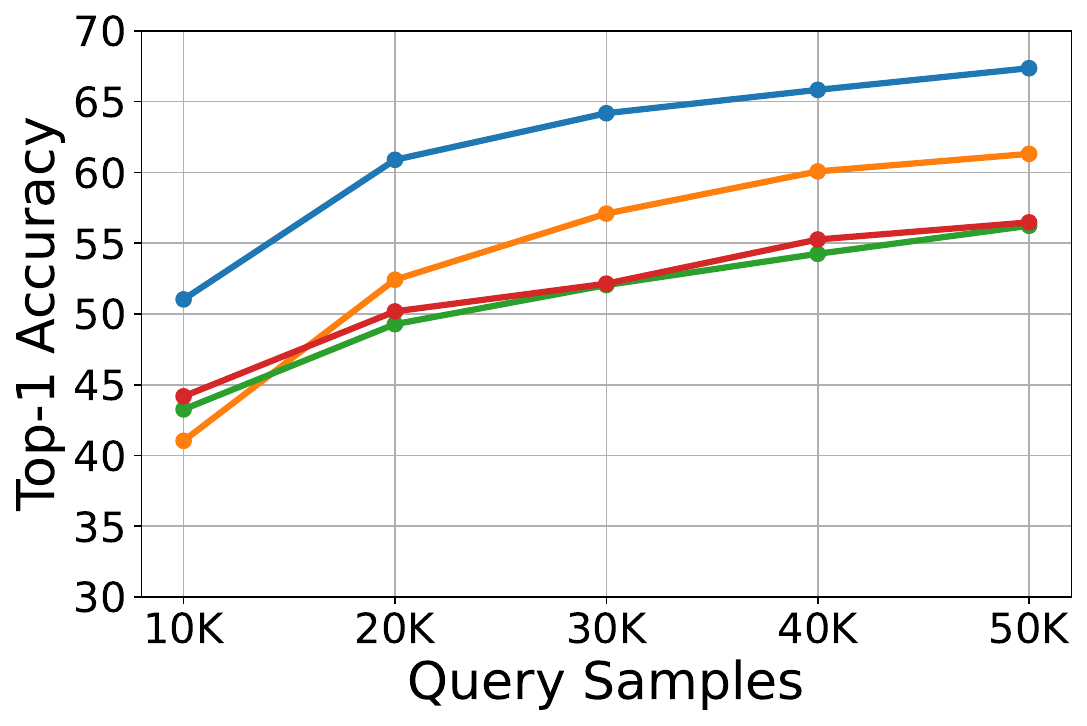}}
    \vspace{-1.5mm}
    \centerline{\hspace{5mm}(b)}
\end{minipage}

\vspace{-1mm}
\centering
\hspace{-2mm}
\begin{minipage}{0.5\linewidth}
    \vspace{3pt}
    \centerline{\includegraphics[width=\textwidth]{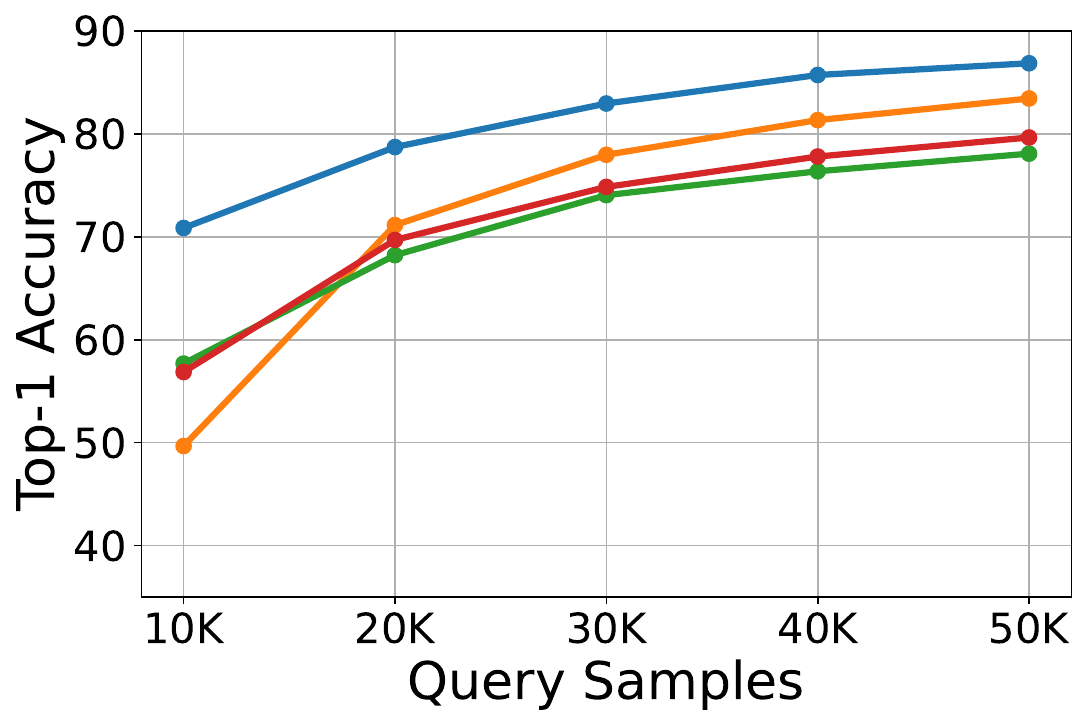}}
    \vspace{-1.5mm}
    \centerline{\hspace{5mm}(c)}
\end{minipage}
\begin{minipage}{0.5\linewidth}
    \vspace{3pt}
    \centerline{\includegraphics[width=\textwidth]{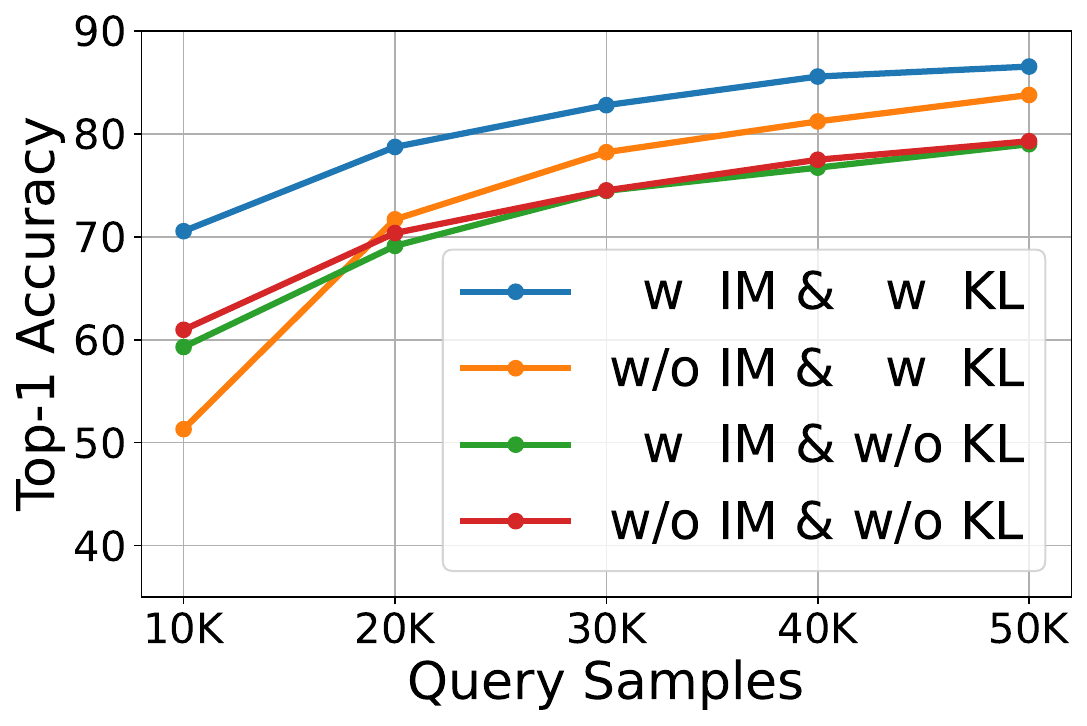}}
    \vspace{-1.5mm}
    \centerline{\hspace{5mm}(d)}
\end{minipage}
\vspace{-3.5mm}
\caption{Curve of top-1 classification accuracy on the datasets of CIFAR-100 (a,b) and CIFAR-10 (c,d). Using MEKD with soft (a,c) or hard (b,d) responses with or without $\mathcal{L}_{IM}$ and $\mathcal{L}_{KL}$. We use ResNet56 as the teacher and use MobileNet as the student.}
\vspace{-6mm}
\label{curve}
\end{figure}

\subsection{Extended Experiments}
\label{extened}
In the real-world application of cloud-to-edge model compression, there are some restrictions, such as the limitation of Internet data exchange and the domain shift in practical scenarios.
We conduct additional experiments to explore the effect of MEKD under these constraints.

\textbf{MEKD with Limited Query Samples.}
\label{limitedquery}
We distill a student MobileNet on CIFAR-10 and CIFAR-100 with a total query sample size ranging from $10K$ to $50K$ with an interval of $10K$.
We report the ACC of MEKD with or without $\mathcal{L}_{IM}$ and $\mathcal{L}_{KL}$.
The curves in Fig.~\ref{curve} show that with more query samples sent to the cloud server, the student model in the edge device can be trained more fully.
We can also analyze from the curves that $\mathcal{L}_{IM}$ does not seem to be that useful without using KLD as an additional distillation loss function, and it gives a big boost to the overall MEKD due to the extra gradient direction of the mapping emulation.

\textbf{MEKD with Out-of-Domain Data.}
We train a teacher (ResNet56 or VGG13) with vanilla supervised learning on Syn. Digits \cite{ganin2016domain}, which contains about $500K$ software-synthesized images. 
We distill a student (MobileNet) on SVHN \cite{netzer2011reading} consisting only of real-shooting photographs.
Tab.~\ref{outofdomain} shows the ACC on the test set of SVHN.
MEKD outperforms most methods in the task of out-of-domain distillation, while DB3KD achieves higher performance due to the use of robust labels \cite{wang2021zero}. 
However, DB3KD leads to a very high data exchange cost between the server and client, since it requires multiple queries to find a mixed image located in the decision boundary to compute robust labels.
In contrast, the data exchange cost of MEKD is much lower.

\renewcommand\arraystretch{1.0}
\begin{table}[t]
\begin{center}
\setlength{\tabcolsep}{1.5mm}{
\scalebox{.5}{}
\begin{tabular}{c|c|c|c}
    \toprule
    \multirow{2}{*}{Teacher} & ResNet56 & VGG13 & \multirow{3}{18mm}{\centerline{Data} \centerline{Exchange}} \\
    & 74.37 & 79.86 & \\
    Student & MobileNet & MobileNet & \\
    \midrule
    KD \cite{hinton2015distilling} & 76.27 & 80.67 & $\sim$ 175 MB \\
    ML \cite{ba2014deep} & 76.78 & 81.90 & $\sim$ 175 MB \\
    AL \cite{wang2018adversarial} & 77.09 & 80.98 & $\sim$ 175 MB \\
    DKD \cite{zhao2022decoupled} & 75.47 & 80.64 & $\sim$ 175 MB \\
    DAFL \cite{chen2019data} & 69.20 & 67.07 & $\sim$ 28.4 GB \\
    KN \cite{orekondy2019knockoff} & 79.65 & 83.37 & $\sim$ 145 MB \\
    AM \cite{wang2020neural} & 84.05 & 86.70 & $\sim$ 11.6 GB \\
    DB3KD \cite{wang2021zero} & 90.15 & 91.14 & $\sim$ 20.8 GB \\
    MEKD (soft) & 86.45 & 88.65 & $\sim$ 120 MB \\
    MEKD (hard) & 86.77 & 89.21 & $\sim$ 120 MB \\
    \bottomrule
\end{tabular}}
\end{center}
\vspace{-5.5mm}
\caption{Top-1 classification accuracy (\%) of methods on SVHN. The teacher models are trained on Syn. Digits with vanilla supervised learning, and achieve the top-1 classification accuracy of 99.56\% for ResNet56 and 99.52\% for VGG13 on Syn.Digits.
}
\label{outofdomain}
\vspace{-6.5mm}
\end{table}

%% file: sec/5_conclusion.tex
\vspace{-1.5mm}
\section{Conclusion}
\label{conclusion}
\vspace{-1mm}

In this paper, we provide a two-step workflow of deprivatization and distillation for B2KD.
Different from aligning logits directly, we theoretically provide a new optimization direction from logits to cell boundaries, and propose a new method of MEKD.
Taking a generator as an inverse mapping of the teacher function does not leak information about the internal structure or parameters of the teacher, because it has a completely different network structure.

\noindent {\bf Limitation.} A well-trained generator is critical in MEKD, and GANs are known to suffer from mode collapse, especially for challenging tasks. 
We alleviate this problem with DCGAN.
Although the parameter size and structural limitations of the model prevent the student from fully mimicking the function of the teacher, MEKD can still improve distillation performance compared with other B2KD methods.


\noindent {\bf Acknowledgement}. This research was supported by Natural Science Fund of Hubei Province (Grant \# 2022CFB823), Alibaba Innovation Research program under Grant Contract \# CRAQ7WHZ11220001-20978282, and HUST Independent Innovation Research Fund (Grant \# 2021XXJS096).